\documentclass[letterpaper]{article} 
\usepackage{aaai23}  
\usepackage{times}  
\usepackage{helvet}  
\usepackage{courier}  
\usepackage[hyphens]{url}  
\usepackage{graphicx} 
\usepackage{natbib}  
\usepackage{caption} 
\usepackage{algorithm}
\usepackage{algorithmic}

\usepackage{newfloat}
\usepackage{listings}
\DeclareCaptionStyle{ruled}{labelfont=normalfont,labelsep=colon,strut=off} 
\floatstyle{ruled}
\newfloat{listing}{tb}{lst}{}
\floatname{listing}{Listing}
\title{AUC Maximization for Low-Resource Named Entity Recognition}

\author{
    Ngoc Dang Nguyen \textsuperscript{\rm 1}, Wei Tan \textsuperscript{\rm 1}, \\
    Lan Du \textsuperscript{\rm 1}\thanks{Corresponding author.}, Wray Buntine \textsuperscript{\rm 2}, Richard Beare \textsuperscript{\rm 1}, Changyou Chen \textsuperscript{\rm 3}
}
\affiliations{
    \textsuperscript{\rm 1}Department of Data Science and Artificial Intelligence, Monash University\\
    \textsuperscript{\rm 2}College of Engineering and Computer Science, VinUniversity\\
    \textsuperscript{\rm 3}Department of Computer Science and Engineering, University at Buffalo\\

}

\usepackage{bibentry}

\usepackage{amsfonts}
\usepackage{amsmath}
\DeclareMathOperator*{\argmax}{arg\,max}
\DeclareMathOperator*{\argmin}{arg\,min}
\usepackage{subeqnarray}
\usepackage{mathtools}

\usepackage{xcolor}

\usepackage{adjustbox}
\usepackage{enumitem}
\usepackage{multirow}
\usepackage{subcaption}
\usepackage{fancyvrb}
\usepackage{comment}

\begin{document}
\maketitle
\begin{abstract}
Current work in named entity recognition (NER) uses either cross entropy (CE) or conditional random fields (CRF) as the objective/loss functions to optimize the underlying NER model. Both of these traditional objective functions for the NER problem generally produce adequate performance when the data distribution is balanced and there are sufficient annotated training examples. But since NER is inherently an imbalanced tagging problem, the model performance under the low-resource settings could suffer using these standard objective functions. Based on  recent advances in area under the ROC curve (AUC) maximization, we propose to optimize the NER model by maximizing the AUC score. We give evidence that by simply combining two binary-classifiers that maximize the AUC score, significant performance improvement over traditional loss functions is achieved under low-resource NER settings. We also conduct extensive experiments to demonstrate the advantages of our method under the low-resource and highly-imbalanced data distribution settings. To the best of our knowledge, this is the first work that brings AUC maximization to the NER setting. Furthermore, we show that our method is agnostic to different types of NER embeddings, models and domains. The code of this work is available at \url{https://github.com/dngu0061/NER-AUC-2T}.
\end{abstract}

\section{Introduction}
Named Entity Recognition (NER) is a fundamental NLP task which aims to locate named entity (NE) mentions and classify them into predefined categories such as location, organization, or person. 
NER usually serves as an important sub-task for
information extraction \cite{ritter2012open},
information retrieval \cite{banerjee2019information}, task oriented dialogues \cite{peng2020soloist}, knowledge base construction \cite{etzioni2005unsupervised} and other language applications. 
Recently, NER has gained significant performance improvements with the advances of pre-trained language models (PLMs) \cite{devlin-etal-2019-bert}. 
Unfortunately, a large amount of training data is often essential for these PLMs to excel and
except for a few high-resource domains, 
the majority of domains ({\it e.g.}, biomedical domain) are scarce and have limited amount of labeled data \cite{10.1093/bioinformatics/btz504}.

\begin{figure}[t]
\centering
\includegraphics[width=\columnwidth,keepaspectratio]{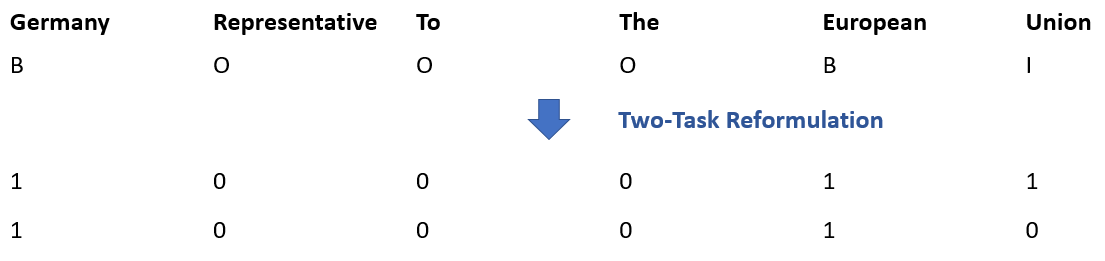} 
\caption{
Reformulating the traditional multi-class NER-tagging problem as a two-task learning problem. 
The first task is to detect if a word is inside an NE, while the second task is to detect if a word is at the beginning of an NE.
}
\label{multitask}
\end{figure}

Since manually annotating a large amount of data in each domain or language is expensive, time-consuming, 
and often infeasible without domain expertise ({\it e.g.,} biomedical NER) \cite{NEURIPS2021_5a7b238b, nguyen-etal-2022-hardness}, 
machine learning practitioners have focused on other alternative approaches to produce effective solution 
to overcome this data scarcity issue. 
These approaches include data augmentation \cite{etzioni2005unsupervised,chen2021data,zhou2022melm}, domain adaptation \cite{10.1145/3366423.3380127}, and multilingual transfers \cite{rahimi2019massively}. 
Although these approaches have shown promising results, they ignore the highly imbalanced data distribution that is inherent to NER corpora. 
Table~\ref{data-description} summarizes this issue. 
While the imbalanced data distribution issue exists in both CoNLL 2003 \cite{tjong-kim-sang-de-meulder-2003-introduction} and OntoNotes5 \cite{weischedel2011ontonotes} where the majority of tag is ``O'', this imbalance issue amplifies given the context of biomedical NER (bioNER). 
For instance, LINNAEUS \cite{article10}, one of the hardest bioNER tasks, 
has nearly 99 percent of its tokens tagged as ``O''. 
Given adequate annotated training examples, 
the NER model could overcome this issue and learn to classify the tokens correctly.
Nonetheless, many corpora, especially those from the biomedical domain, can be scarcely labeled. 

Since NER corpora can be both scarcely labeled and highly imbalanced, 
we argue that the standard cross entropy objective function, though capable of producing the optimal token classifier asymptotically under the maximum likelihood principle, would fail to achieve the desirable performance under the low-resource and imbalanced settings. 

\begin{table*}[t]
\setlength{\tabcolsep}{4.5pt}
\centering
\small
\begin{adjustbox}{max width=\textwidth}
\begin{tabular}{|l|ccc|ccc|ccc|ccc|ccc|}
    \hline

        \multicolumn{1}{|l|}{\bf Dataset}
        & \multicolumn{3}{c|}{\bf \# sentences}
        & \multicolumn{3}{c|}{\bf \# tokens}
        & \multicolumn{9}{c|}{\bf \% label (B/I/O)}
    \\ \cline{2-16}
    
    \multicolumn{1}{|l|}{}
        & {\bf Train} & {\bf Dev} & {\bf Test}
        & {\bf Train} & {\bf Dev} & {\bf Test}
        & \multicolumn{3}{c|}{\bf Train} & \multicolumn{3}{c|}{\bf Dev} & \multicolumn{3}{c|}{\bf Test}
    \\ \hline 
    
     \multicolumn{1}{|l|}{CoNLL 2003~\shortcite{tjong-kim-sang-de-meulder-2003-introduction}}
        & {14,987} & {3,466} & {3,684}
        & {203,621} & {51,362} & {46,435}
        & {11.5}&{5.2}&{83.3} & {11.5}&{5.2}&{83.3} & {12.2}&{5.3}&{82.5}
    \\ \hline 
    
     \multicolumn{1}{|l|}{OntoNotes5~\shortcite{weischedel2011ontonotes}}
        & {20,000} & {3,000} & {3,000}
        & {364,611} & {54,423} & {55,827}
              & {6.2}&{4.9}&{89.1} & {6.2}&{4.9}&{89.1} & {6.1}&{4.9}&{89.0}
    \\ \hline 
    
     \multicolumn{1}{|l|}{LINNAEUS~\shortcite{article10}}
        & {11,934} & {4,077} & {7,141}
        & {281,273} & {93,877} & {165,095}
              & {0.7}&{0.4}&{98.9} & {0.8}&{0.4}&{98.8} & {0.9}&{0.5}&{98.6}
    \\ \hline 
    
    \multicolumn{1}{|l|}{NCBI~\shortcite{10.5555/2598938.2599127}}
        & {5,423} & {922} & {939}
        & {135,701} & {23,969} & {24,497}
              & {3.8}&{4.5}&{91.7} & {3.3}&{4.5}&{92.2} & {3.9}&{4.4}&{91.7}
    \\ \hline 
    
     \multicolumn{1}{|l|}{s800~\shortcite{10.1371/journal.pone.0065390}}
        & {5,732} & {829} & {1,629}
        & {147,291} & {22,217} & {42,298}
              & {1.7}&{2.2}&{96.1} & {1.7}&{2.2}&{96.1} & {1.8}&{2.5}&{95.7}
    \\ \hline 
    
    \hline    
    \end{tabular}
\end{adjustbox}
\caption{
The size of the benchmark NER/bioNER corpora as well as the label distribution for each corpus. In this work, we use the well-defined BIO tagging scheme, {\it i.e.}, B-Begin, I-Inside, and O-Outside of NEs.
}
\label{data-description}
\end{table*}

With the above desiderata, 
this paper considers optimizing the NER model under a different objective/loss function. 
Instead of optimizing the NER model using the standard cross entropy loss, 
we propose to optimize the model using the area under ROC curve (AUC) score. 
Direct optimization of the AUC score has been shown to greatly benefit the performance of highly imbalanced tasks since maximizing AUC aims to rank the prediction score of any positive data higher than that of any negative data \cite{ying2016stochastic,yuan2020large}. 
However, most recent robust/practical AUC maximization techniques are solely designed to solve for the binary classification task. 
Thus, this makes direct AUC optimization implementation infeasible for NER. 
Consequently, we propose to turn the standard BIO tagging problem into a two-task problem. 
Figure~\ref{multitask} shows how the standard BIO tagging scheme can be converted into two separate tasks, each of which can be optimized with AUC score. 
At prediction time, the predictions of the two tasks can be combined to generate the final prediction that matches the output normally expected by the BIO tagging scheme.

To demonstrate the effectiveness of our proposed method, 
we conduct extensive empirical studies on the corpora listed in Table~\ref{data-description}. 
The corpora are from both generic domains, {\it e.g.}, CoNLL 2003 and OntoNotes5, and specialized domains, {\it e.g.}, LINNAEUS, NCBI and s800.
Using corpora from different domains would substantiate that our method is domain agnostic.
Additionally, employing different model architecture such as transformers BERT \cite{devlin-etal-2019-bert} and seq2seq BART \cite{lewis2019bart} would indicate that our method is model agnostic.
Lastly, by experimenting with different pre-trained embeddings, 
we give evidence that our method is embedding agnostic.
Our studies reveals that our method exhibits significant performance improvement over 
the standard objective/loss functions by a large margin. 

We summarize our contributions as follows:
\begin{itemize}
    \item Two-task NER reformulation: We reformulate the standard NER multi-class approach as a two-task learning problem. Under this new setting, two binary classifiers are used to predict respectively if a word is inside an NE and if a word is at the start of an NE. This simple reformulation makes the AUC optimization feasible for NER. 
    \item A new AUC loss for NER: Instead of optimising the standard loss functions, we adapt the idea of AUC maximization to tackle the lower-resource and imbalanced NER.
    \item Comprehensive experiments: We conduct extensive empirical studies of our method on a broad range of settings, and demonstrate its consistently strong performance compared with the standard objective functions.
\end{itemize}

\section{Backgrounds}
\subsection{Named Entity Recognition}
NER is a key task in NLP systems such as question-answering, information-retrieval, co-reference resolution, and topic modelling \cite{yadav2019survey}. 
Although there are many variations regarding the definition of named entity (NE), 
the following types remain prevalent 
1) generic NEs ({\it e.g.}, person, organization, and location), 
and 2) domain-specific NEs ({\it e.g.}, virus, protein, and genome) \cite{DBLP:journals/corr/abs-1812-09449,Kripke1980-KRINAN}. 
For instance, the NER task includes but is not limited to the identification of person, location, or organization in either short or long unstructured texts. 
In biomedical domains, bioNER is to properly classify virus/disease names into predefined categories. 
Given a set of tokens $\mathbf{x} = \{x_1,\ldots,x_l\}$, the algorithm would output a list of appropriate tags $\mathbf{y} = \{y_1,\ldots,y_l\}$ \cite{DBLP:journals/corr/abs-1812-09449}. NER is considered a challenging task for two reasons \cite{DBLP:journals/corr/LampleBSKD16}: 1) in most languages and domains, the amount of manually labeled training data for NER is limited; and  2) it is difficult to generalize from a small sample of training data due to the inherent imbalance nature of NER.

\subsection{AUC Maximization}
AUC (Area Under ROC Curve) has been used in various machine learning works as an important measuring criterion \cite{freund2003efficient,kotlowski2011bipartite,zuva2012evaluation}. 
Since AUC is non-convex, discontinuous and sensitive to model change,
direct optimization of AUC score often leads to an NP-hard problem \cite{yuan2020large}. Thus, many works have tried to alleviate the computational difficulties by providing solutions through a pairwise surrogate loss \cite{freund2003efficient}, a hinge loss \cite{zhao2011online}, and a least square loss \cite{gao2013one}.
\citet{yuan2020large} pointed out that although the proposed least-square surrogate loss \cite{gao2013one,liu2019stochastic} makes AUC maximization scalable, it has two largely ignored issues which are 1) it has an adverse effect when trained with well-classified data ({\it i.e.,} easy data), and 2) it is sensitive to noisily labeled data ({\it i.e.,} noisy data). Thus, they proposed to decompose this surrogate least-square loss by what they call AUC margin loss and achieved the 1st place on Stanford CheXpert competition. 
Further improvements to AUC optimization have been observed via compositional training \cite{yuan2021compositional}, {\it i.e.}, alternating between the cross entropy loss and the AUC surrogate loss during training.
AUC maximization works well when 1) there exists an imbalanced data distribution, 
{\it e.g.}, classification of chest x-ray images to identify rare threatening diseases, or classification of mammogram for breast cancer study \cite{yuan2020large}; 
or 2) the AUC score is the default metric for evaluating and comparing different methods, {\it i.e.}, directly maximizing AUC score can lead to large improvement in the model performance \cite{yuan2020large}.

\section{AUX Maximization for NER}
\subsection{Notation}
We will use common notation found in the AUC maximization literature \cite{zhao2011online,yuan2021compositional,yuan2020large,yang2021learning} to present the problem of AUC maximization in the context of the NER task. 
Let $\mathbb{I}(\cdot)$ be an indicator function of a predicate, $[s]_{+}=\max (s, 0)$. 
We denote $\mathcal{S}=\left\{\left(\mathbf{x}_{1}, \mathbf{y}_{1}\right), \ldots,\left(\mathbf{x}_{n}, \mathbf{y}_{n}\right)\right\}$ as a set of training data, 
where $\mathbf{x}_{i} = {x}_i^1, \ldots, {x}_i^l$ represents the $i$-th training example  ({\it i.e.}, a sentence of length $l$), while $\mathbf{y}_{i} \in\{\text{B}, \text{I}, \text{O}\}^{l}$ denotes its corresponding sequence of labels.
Additionally, $\mathbf{z}=({x}, {y})$ will also be used for ease of reference. Following standard machine learning convention, we use $\mathbf{w} \in \mathbb{R}^{d}$ to represent the parameters of the deep neural network to be learned, and $h_{\mathbf{w}}(\mathbf{x})=h(\mathbf{w}, \mathbf{x})$ to denote the objective mapping function $h:\mathcal{X}\rightarrow\mathcal{Y}$. 
Lastly, the standard approach of deep learning is to define a loss/objective function on individual data by $L(\mathbf{w} ; \mathbf{x}, y)=\ell\left(h_{\mathbf{w}}(\mathbf{x}), \mathbf{y}\right)$, 
where $\ell(\hat{\mathbf{y}}, \mathbf{y})$ is a surrogate loss function of the mis-classification error ({\it e.g.}, cross-entropy loss, conditional random fields), 
and to minimize the empirical loss $\min _{\mathbf{w} \in \mathbb{R}^{d}} \frac{1}{n} \sum_{i=1}^{n} L\left(\mathbf{w} ; \mathbf{x}_{i}, \mathbf{y}_{i}\right)$. 
However, as we previously discussed, this standard surrogate loss function is not suitable for situations where there exists an imbalanced data distribution. 
This imbalanced data distribution issue is the inherent problem for the NER/bioNER task as most corpora have the majority of labels tagged as ``O'' (see Table~\ref{data-description}).

\subsection{Reformulation of NER}
Existing work on AUC maximization typically follows the Wilcoxon-Mann-Whitney statistic \cite{hanley1982meaning,clemenccon2008ranking,yuan2020large} and interprets the $\mathrm{AUC}$ score as the probability of a positive sample ranking higher than a negative sample:
{\scriptsize
    \begin{subeqnarray}
        \lefteqn{\operatorname{AUC}(\mathbf{w})}\nonumber\\
        &=& \; \operatorname{Pr}\left(h_{\mathbf{w}}({x}) \geq h_{\mathbf{w}}\left({x}^{\prime}\right) \mid y=1, y^{\prime}=-1\right)\\
        &=& \; \mathbb{E}\left[\mathbb{I}\left(h_{\mathbf{w}}({x})-h_{\mathbf{w}}\left({x}^{\prime}\right) \geq 0\right) \mid y=1, y^{\prime}=-1\right].
    \label{eq:AAAIAUCDefinition}
    \end{subeqnarray}
}

\noindent With $N_{+}$, and $N_{-}$ denoting the size of positive and negative training data set respectively, many existing works formulate the AUC maximization on training data $\mathcal{S}\equiv\{\mathcal{S}_{-},\mathcal{S}_{+}\}$ as
{\scriptsize
    \begin{subeqnarray}
        \lefteqn{\argmax_{\mathbf{w} \in \mathbb{R}^{d}} \operatorname{AUC}(\mathbf{w})}\nonumber\\
        &=&\argmin_{\mathbf{w} \in \mathbb{R}^{d}} \operatorname{Pr}\left(h_{\mathbf{w}}({x}) < h_{\mathbf{w}}\left({x}^{\prime}\right) \mid y=1, y^{\prime}=-1\right)\\
        &=& \argmin_{\mathbf{w} \in \mathbb{R}^{d}}\mathbb{E}\left[\mathbb{I}\left(h_{\mathbf{w}}({x})-h_{\mathbf{w}}\left({x}^{\prime}\right) < 0\right) \mid y=1, y^{\prime}=-1\right]\\
        &\equiv&  
        \argmin_{\mathbf{w} \in \mathbb{R}^{d}} \frac{1}{N_{+} N_{-}} \sum_{{x} \in \mathcal{S}_{+}} \sum_{{x}^{\prime} \in \mathcal{S}_{-}} \mathbb{I}\left(h_{\mathbf{w}}({x})-h_{\mathbf{w}}\left({x}^{\prime}\right) < 0\right).
        \label{eq:AAAIAUCLoss}
    \end{subeqnarray}
}

\noindent While the definition from Eqs~\eqref{eq:AAAIAUCDefinition} and \eqref{eq:AAAIAUCLoss} subsequently leads to the development of many prominent works in AUC maximization, it is not directly applicable to the NER problem.

\begin{figure}[t]
\centering
\includegraphics[width=\columnwidth,keepaspectratio]{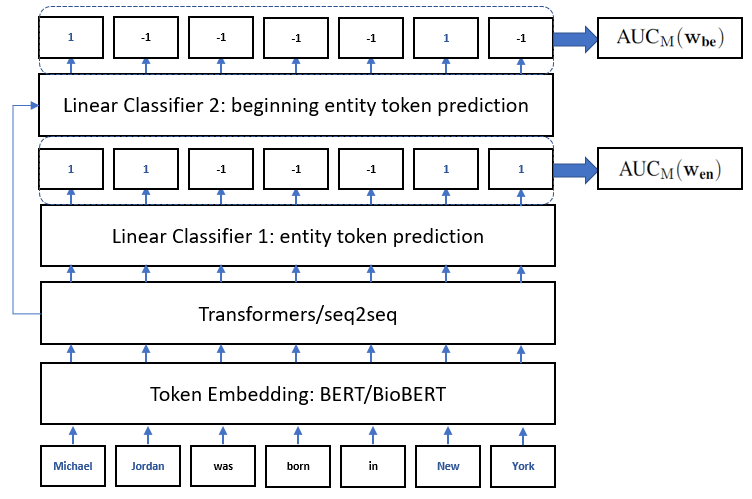} 
\caption{
Reformulating the traditional multi-class NER-tagging problem as a two-task problem to be optimized with AUC.
The embedding and model parameters are shared while the linear classifiers have their own parameter set.
}
\label{AUCLossMT}
\end{figure}

To make direct AUC maximization applicable to NER, we propose to reformulate the standard NER multi-class setting into a two-task setting as shown in Figure~\ref{AUCLossMT}. 
In this setting, the standard BIO-tagging scheme is broken into two parallel tasks. 
One of the tasks is to detect if the token $x_i^l$ belongs to a named entity, we name the label of this task $\mathbf{y}_{\textbf{en}_i}\in\{-1,1\}^{l}$. 
The other task is to detect if the token $x_i^l$ is the beginning token of the named entity, we name the label of this task $\mathbf{y}_{\textbf{be}_i}\in\{-1,1\}^{l}$, {\it i.e.}, $\mathbf{y}_{i} = \{ \mathbf{y}_{\textbf{en}_i},\mathbf{y}_{\textbf{be}_i} \}$.
Based on this two-task setting, the parameters set $\mathbf{w}$ can be separated into two sets, $\mathbf{w}_{\textbf{en}} = \{\theta, \omega_{\textbf{en}}\}$ and $\mathbf{w}_{\textbf{be}} = \{\theta, \omega_{\textbf{be}}\}$.
$\theta$ denotes the shared embedding and language model parameters, while $\omega_{\textbf{en}}$ and $\omega_{\textbf{be}}$ denote the parameters for the entity-token classifier and the beginning-token classifier respectively.
By reformulating the standard NER setting into two separate tasks, we can maximize the AUC score of each task.
Since the two tasks inherit the imbalanced data distribution problem from the NER multi-class setting, we expect maximizing the AUC score will lead to a better classifier for each task, and eventually improve the NER performance.

\subsection{AUC Maximization for NER Two-Task Setting}
Although there are many AUC maximization techniques that can be applied to this NER two-task setting, we choose the deep AUC margin loss (DAM) \cite{yuan2020large} due to its 1) robustness/scalability of implementation to deep neural networks, and 2) ability to overcome the critics from which previous AUC maximization approaches such as the least-squared surrogate loss might suffer \cite{ying2016stochastic,natole2018stochastic,liu2018fast,liu2019stochastic,yuan2020large}. Under DAM, we define the objective/loss function to the entity-token prediction task as follows:
{\scriptsize
    \begin{subeqnarray}
        \lefteqn{\text{AUC}_{\mathrm{M}}(\mathbf{w}_{\textbf{en}})}\nonumber\allowdisplaybreaks
        \\&=&
        \mathbb{E}\left[\left(m-h_{\mathbf{w}_{\textbf{en}}}({x})+h_{\mathbf{w}_{\textbf{en}}}\left({x}^{\prime}\right)\right)^{2} \mid y=1, y^{\prime}=-1\right]\allowdisplaybreaks
        \\&=&
        \min _{a, b}{A_{1}(\mathbf{w}_{\textbf{en}})} 
        +{A_{2}(\mathbf{w}_{\textbf{en}})}
        +
        (m-a+b)^{2}\allowdisplaybreaks
        \\&=&
        \min _{a, b}A_{1}(\mathbf{w}_{\textbf{en}})+A_{2}(\mathbf{w}_{\textbf{en}})+\max _{\alpha\geq0}\left\{2 \alpha(m-a
        +b
        )
        -\alpha^{2}\right\},\allowdisplaybreaks
    \label{eq:AAAIAUCMargin}
    \end{subeqnarray}
}

\noindent where $A_{1}(\mathbf{w}_{\textbf{en}})=\mathbb{E}[h^2_{\mathbf{w}_{\textbf{en}}}({x})\mid y=1] - a^2$, and $A_{2}(\mathbf{w}_{\textbf{en}})=\mathbb{E}[h^2_{\mathbf{w}_{\textbf{en}}}({x})\mid y=-1] - b^2$. 
Using the definition for the variance, the minimization problem of $a$ and $b$ is achieved when $a=a(\mathbf{w}_{\textbf{en}})=\mathbb{E}[h_{\mathbf{w}_{\textbf{en}}}({x})\mid y=1]$, and $b=b(\mathbf{w}_{\textbf{en}})=\mathbb{E}[h_{\mathbf{w}_{\textbf{en}}}({x}^{\prime})\mid y=-1]$ respectively. 
Examining Eq~\eqref{eq:AAAIAUCMargin} reveals that maximizing AUC requires the predictor 1) to have low variance on both negative and positive words, and 2) to push the mean prediction scores of positive and negative words to be far away based on the value of the margin $m$.
We expect that minimizing Eq~\eqref{eq:AAAIAUCMargin} with respect to $\mathbf{w}_{\textbf{en}}$ can lead to a reliable predictor for the imbalanced entity-token prediction task.
Following the same steps, we can define a similar loss function for the beginning-token prediction task. Since this is similar, we will skip this definition in the paper. Lastly, we adhere to the standard multi-task setting and optimize for both tasks by minimizing for the following loss:
{
    \begin{equation}
        \text{AUC}_{\mathrm{M}}(\mathbf{w}_{\textbf{en}}) + \lambda\text{AUC}_{\mathrm{M}}(\mathbf{w}_{\textbf{be}}),
    \label{eq:AAAIAUCBothLoss}
    \end{equation}
}

\noindent with $\lambda$ controlling the trade-off between the two losses.

\begin{algorithm}[tb]
\caption{NER two-task prediction}
\label{alg:MTPrediction}
\textbf{Input}: $\mathcal{S}_{\text{test}}$, $\mathbf{w}^{*}_{\textbf{en}}$, $\mathbf{w}^{*}_{\textbf{be}}$\\
\textbf{Output}: {\normalsize $\hat{\mathbf{y}}_1, \ldots, \hat{\mathbf{y}}_{\text{n-test}}$, s.t., $\hat{\mathbf{y}}_i \in\{\text{B, I, O}\}^{l}$}\\
\begin{algorithmic}[1] 
\FOR{$\mathbf{x}_i \in \mathcal{S}_{\text{test}}$}
\STATE $\hat{\mathbf{y}}_{\textbf{en}_i} = h_{\mathbf{w}^{*}_{\textbf{en}}}(\mathbf{x}_i)$
\STATE $\hat{\mathbf{y}}_{\textbf{be}_i} = h_{\mathbf{w}^{*}_{\textbf{be}}}(\mathbf{x}_i)$
\FOR{$j$ in {\tt range}(l)}
\IF {$\hat{{y}}_{\text{en}_i}^j \equiv 1$ and $\hat{{y}}_{\text{be}_i}^j \equiv 1$}
\STATE $\hat{{y}}_i^j :=$ ``B''
\ELSIF{$\hat{{y}}_{\text{en}_i}^j \equiv 1$ and $\hat{{y}}_{\text{be}_i}^j \equiv -1$}
\STATE $\hat{{y}}_i^j :=$ ``I''
\ELSE
\STATE $\hat{{y}}_i^j :=$ ``O''
\ENDIF
\ENDFOR
\ENDFOR
\STATE \textbf{return} {\normalsize $\hat{\mathbf{y}}_1, \ldots, \hat{\mathbf{y}}_{\text{n-test}}$, s.t., $\hat{\mathbf{y}}_i \in\{\text{B, I, O}\}^{l}$}
\end{algorithmic}
\end{algorithm}

\subsection{NER Two-Task Prediction}
At prediction time, we follow Algorithm~\ref{alg:MTPrediction} to generate the prediction tags expected by the standard BIO-tagging scheme. 
$\mathbf{w}^{*}_{\textbf{en}}$, and $\mathbf{w}^{*}_{\textbf{be}}$ represent the optimal parameters learnt by minimizing Eq~\eqref{eq:AAAIAUCBothLoss} during training.
After receiving the predictions from the entity-token and the beginning-token prediction tasks, we linearly combine both sets of predictions to generate the final BIO tags as demonstrated in Algorithm~\ref{alg:MTPrediction}. 
We acknowledge that this linear combination of predictions might raise a certain inconsistency during prediction time when $\hat{{y}}_{\text{en}_i}^j \equiv -1$ and $\hat{{y}}_{\text{be}_i}^j \equiv 1$. 
Currently, we treat the prediction when this happens as ``O''.
Our justification includes 
1) ``O'' is the major tag for most of the NER corpora (Table~\ref{data-description}); thus, predicting ``O'' would align with the data statistics, 
and 
2) all of the current standard evaluation metrics for NER do not take the true negative prediction into the calculation for the performance score; 
thus, predicting ``O'' would lead to no improvement in the 
evaluation score.

\section{Experimental Settings}
We verify the performance of our AUC NER two-task reformulation on both the imbalanced and the low-resource NER settings.
All of the experiments were set up as follows.

\subsection{Domains and Corpora}
We used corpora from both the general domain and the biomedical domain. 
Table~\ref{data-description} summarizes the statistics for these corpora. 
Both CoNLL 2003 \cite{tjong-kim-sang-de-meulder-2003-introduction} and OntoNotes5 \cite{weischedel2011ontonotes} are standard corpora from the general domain to evaluate and benchmark the NER performance.
Whereas NCBI \cite{10.5555/2598938.2599127}, LINNAEUS \cite{article10}, and s800 \cite{10.1371/journal.pone.0065390} are standard NER corpora used for biomedical named entity recognition.
While NCBI is often used to train the NER model to identify the disease-NEs, both LINNAEUS and s800 are trained to detect species-NEs.
Thus, they exhibit different linguistic characteristics.
We chose these biomedical corpora to verify the performance of our method since:
\begin{itemize}
    \item The data distribution of biomedical corpora is often heavily imbalanced, 
    which make them the perfect candidates for AUC maximization in NER. 
    All included biomedical corpora have higher percentage of ``O'' tag compared to those corpora from the general domain (see Table~\ref{data-description}).
    \item Standard approaches of transferring PLMs from high-resource domains to low-resource ones often fail to achieve satisfactory performance for biomedical corpora \cite{10.1093/bioinformatics/btz504}. This happens as clinical narratives can contain challenging linguistic characteristics which do not overlap with those in other domains \cite{Patel2005ThinkingAR}, as also observed in our medical NLP project.
\end{itemize}
\noindent By demonstrating the significance of our method over the baselines for both general and specialized domains, it substantiates that our method is both domain and data agnostic.

\begin{figure*}[t]
\centering
    \begin{subfigure}[t]{0.5\textwidth}
        \centering
        \includegraphics[width=\textwidth]{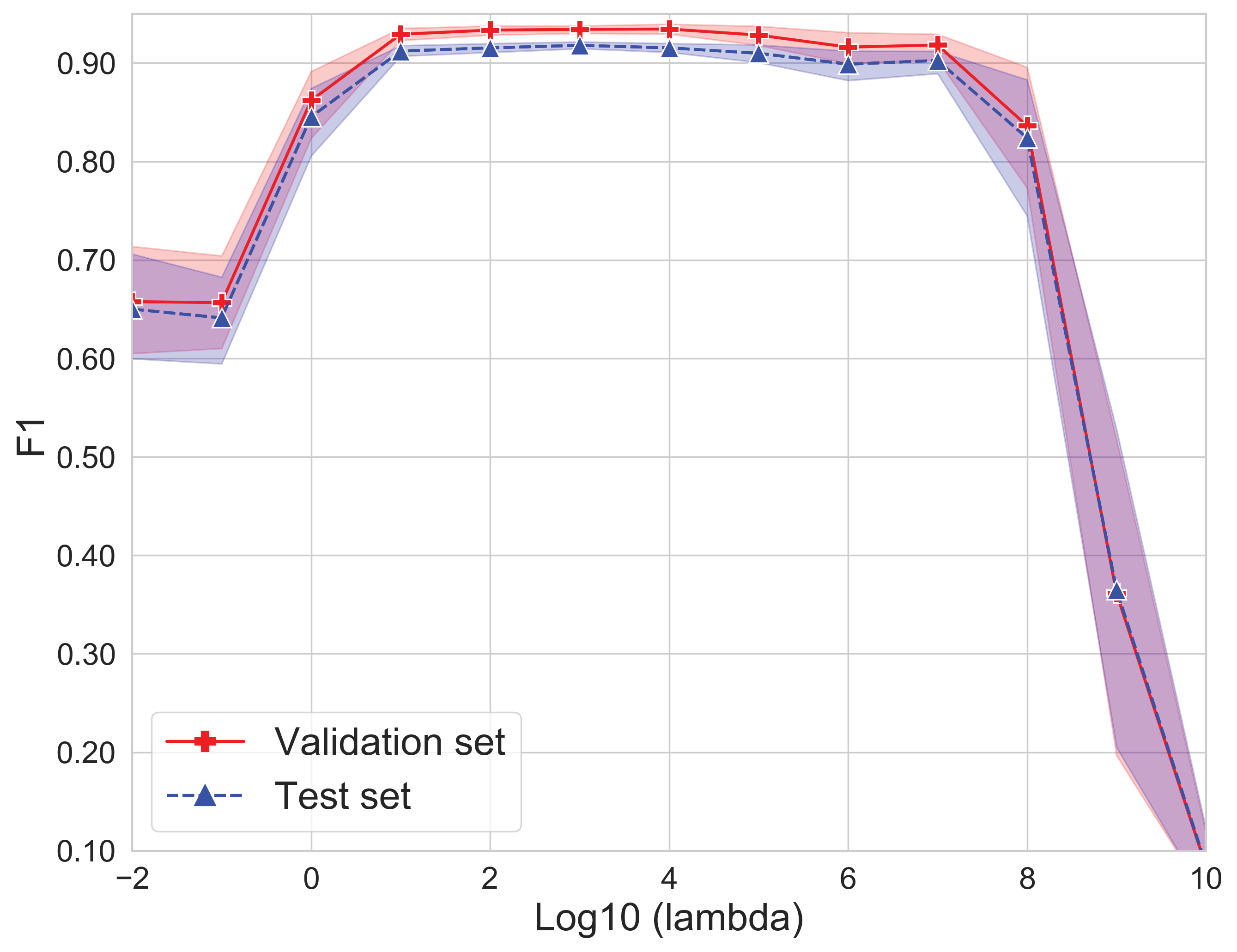}
        \caption{AUC-2T performance}
        \label{fig:ablation_lamb_auc}
    \end{subfigure}%
    ~ 
    \begin{subfigure}[t]{0.5\textwidth}
        \centering
        \includegraphics[width=\textwidth]{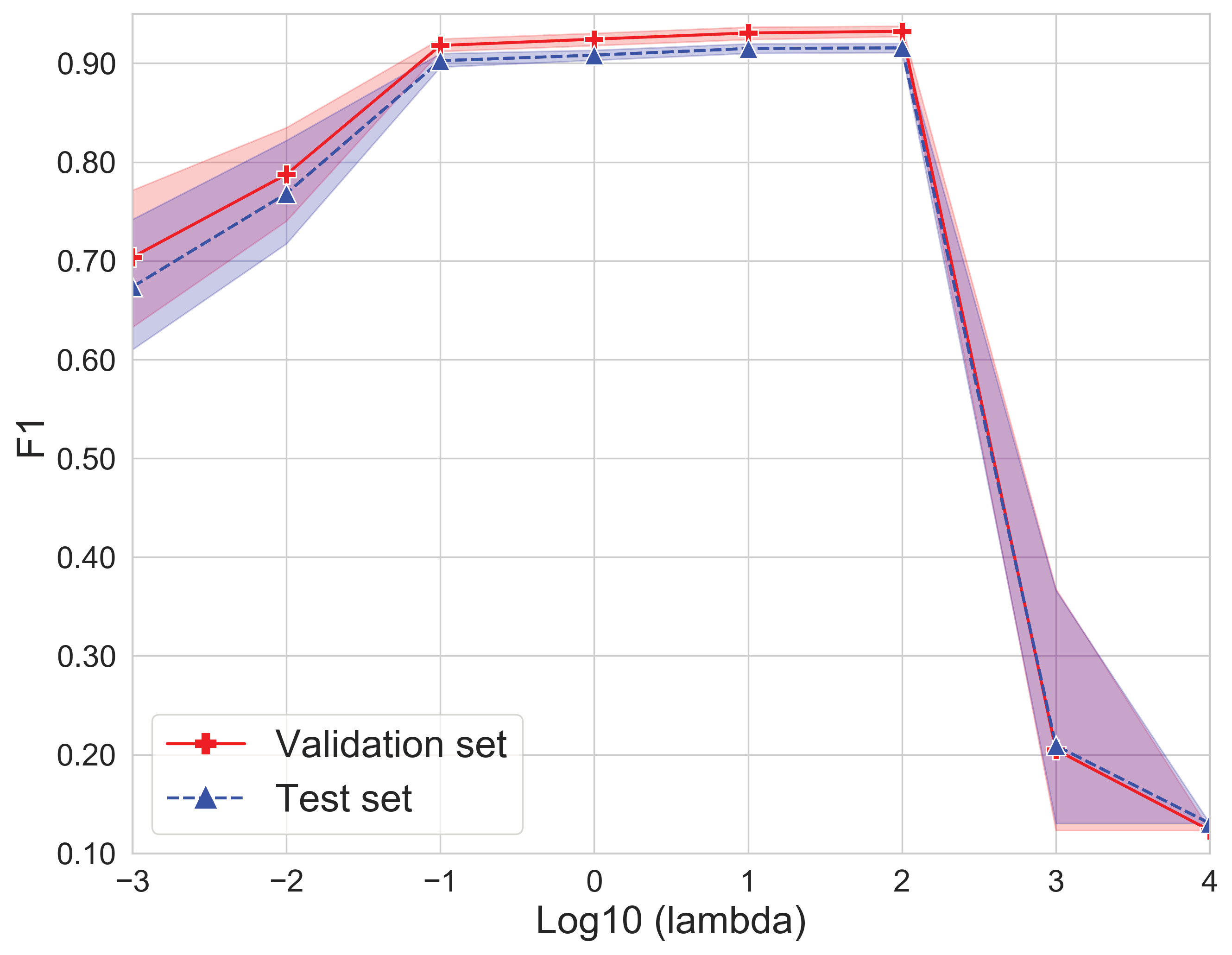}
        \caption{COMAUC-2T performance}
         \label{fig:ablation_lamb_comauc}
    \end{subfigure}
    \caption{Average model performance taken from 10 random partitions of 200 CoNLL 2003 sentences. We plot F1-scores as a function of the logarithm of $\lambda$ values. In both plots, the red line represents the F1-scores based on the validation set and the blue dash line represents the F1-scores based on the test set. The error bands indicate the 95\% confidence level of the F1-scores.
    }
    \label{fig:ablation_lamb}
\end{figure*}

\subsection{Model Architecture and Embedding}
For embeddings and model architectures, 
we used both ``bert-base-cased'' \cite{devlin-etal-2019-bert} and ``facebook/bart-base'' \cite{lewis2019bart} 
on CoNLL 2003 and OntoNotes5. 
``bert-base-cased'' uses the transformers model architecture, while ``facebook/bart-base'' uses the seq2seq model architecture.
By testing our AUC NER two-task method on different embeddings and model architectures, we give evidence that our method is also embedding and model agnostic.
Due to their distinct linguistic characteristics which often lead to the out-of-vocabulary (OOV) issue when training with the standard language model embeddings, all biomedical corpora will be trained with ``biobert-base-cased-v1.1'' \cite{10.1093/bioinformatics/btz682}. 
The implementation of BioBERT, which achieved the state-of-the-art performance for many benchmark biomedical corpora, is publicly available at BioBERT github website (see {\tt \url{https://github.com/dmis-lab/biobert}} \cite{10.1093/bioinformatics/btz682}).

\subsection{Low-Resource and Imbalanced Settings}
To analyze the performance of our AUC NER two-task method under the low-resource and imbalanced distribution scenarios, we consider the following experimental settings:
\begin{itemize}
    \item The size of training set $\mathcal{S}$: We used $\mathcal{S}^{}$ with size $\in \{20,  $ $ 50, 100,150, 200, 250, 300, 350, 400, 450, 500, 1000\}$ to simulate the data scarcity issue in low-resource scenarios. 
    Given the size of $\mathcal{S}$, we trained our method and all the baselines on multiple training partitions of the same size and report/investigate their average performance.
    \item We developed an imbalanced data generator for the NER task. 
    This generator samples $\mathcal{S}$ that contains $\{1, 2, 5, 10, 20\}$ percentage of entity-tokens. 
    By evaluating the performance of our method on different data distributions, we can indicate the robustness of our method under different imbalanced data distribution setups. 
\end{itemize}

\subsection{Baselines \& AUC NER Two-Task Method}
For our baselines, we chose the following methods:
\begin{itemize}
    \item \textbf{CE}: The standard cross entropy loss, commonly used in NER, was used as one of the major baselines to verify the significance/impact of our method. We used all standard hyperparameter settings to record its performance.
    \item \textbf{CRF}: CRF 
    represents our second major baseline as it has been traditionally used in many NER works, such as those of \citet{huang2015bidirectional, DBLP:journals/corr/LampleBSKD16,  10.1007/978-3-319-64861-3_33, XU2019122}. It should be noted that as CRF uses the transition matrix to generate the predictions, we cannot use the WordPiece tokenization for this baseline.
    \item \textbf{CE-2T}: This is our last baseline. 
    For this baseline, we broke the standard multi-class NER setting into the two-task NER setting as shown in both Figure~\ref{multitask} and Figure~\ref{AUCLossMT}. 
    However, instead of optimizing both tasks under the AUC maximization approach, 
    we used the binary cross entropy loss. 
    This baseline serves to demonstrate that the significance of our work is related not only to the NER two-task setting but also to the AUC objective function.
\end{itemize}
For our AUC NER two-task method, we considered:
\begin{itemize}
    \item \textbf{AUC-2T}: AUC-2T, one of the two implementations to verify the significance/impact of our work, learns the optimal $\mathbf{w}^{*}_{\textbf{en}}$, and $\mathbf{w}^{*}_{\textbf{be}}$ by minimizing Eq~\eqref{eq:AAAIAUCBothLoss} during training.
    \item \textbf{COMAUC-2T}: COMAUC-2T also minimizes Eq~\eqref{eq:AAAIAUCBothLoss} at training time. However, different from AUC-2T, COMAUC-2T learns $\mathbf{w}^{*}_{\textbf{en}}$ and $\mathbf{w}^{*}_{\textbf{be}}$ by following the idea proposed by \citet{yuan2021compositional} and alternates between the standard multi-class cross entropy loss function and the  AUC two-task loss function during training\footnote{Since CRF uses whole word instead of sub-tokens, we cannot alternate the training between CRF and AUC optimization.}.
\end{itemize}
Both our AUC-2T and COMAUC-2T were set up for the NER task by modifying the code from {\tt libauc} \cite{libauc2022}. At prediction time, both AUC-2T and COMAUC-2T use Algorithm~\ref{alg:MTPrediction} to generate the BIO-tag for evaluation.

\subsection{Evaluation Metrics}
Since there is no agreed upon AUC evaluation metric for the NER task, we used the standard F1, precision and recall score 
to evaluate for all our setups \cite{seqeval}.

\section{Experimental Results \& Discussions}
\begin{table*}[t]
\setlength{\tabcolsep}{.5pt}
\centering
\small
\begin{adjustbox}{max width=\textwidth}
\begin{tabular}{|l|ccc|ccc|ccc|}
    \hline
        \multicolumn{1}{|l|}{\bf Training Size}
        & \multicolumn{3}{c|}{\bf 20}
        & \multicolumn{3}{c|}{\bf 50}
        & \multicolumn{3}{c|}{\bf 100}
    \\ \cline{2-10}

    \multicolumn{1}{|c|}{}
        & {\bf Precision} & {\bf Recall} & {\bf F1}
        & {\bf Precision} & {\bf Recall} & {\bf F1}
        & {\bf Precision} & {\bf Recall} & {\bf F1}
    \\ \hline

    \textbf{CE}&        
        ${0.2889}_{0.0286}$&${0.2425}_{0.0282}$&${0.2189}_{0.0214}$&${0.4451}_{0.0243}$&${0.4011}_{0.0377}$&${0.3921}_{0.0285}$&${0.5953}_{0.0210}$&${0.6683}_{0.0225}$&${0.6253}_{0.0206}$\\
       
    \textbf{CRF}
        &${0.5120}_{0.0156}$&${0.5910}_{0.0228}$&${0.5437}_{0.0177}$&${0.6120}_{0.0125}$&${0.6592}_{0.0129}$&${0.6318}_{0.0108}$&${0.6731}_{0.0150}$&${0.6944}_{0.0122}$&${0.6817}_{0.0124}$\\
    
    \textbf{CE-2T}
        &${0.0783}_{0.0001}$&${0.3915}_{0.0001}$&${0.1304}_{0.0001}$&${0.0783}_{0.0001}$&${0.3915}_{0.0001}$&${0.1304}_{0.0001}$&${0.0783}_{0.0001}$&${0.3915}_{0.0001}$&${0.1304}_{0.0001}$\\
        
    \textbf{AUC-2T}
        &$\underline{0.5794}_{0.0264}$&$\underline{0.7000}_{0.0224}$&$\underline{0.6289}_{0.0241}$&$\underline{0.7976}_{0.0174}$&$\underline{0.8693}_{0.0076}$&$\underline{0.8300}_{0.0131}$&$\underline{0.8589}_{0.0139}$&$\underline{0.8973}_{0.0122}$&$\underline{0.8774}_{0.0131}$\\
    
    \textbf{COMAUC-2T}
        &${\bf 0.6647}_{0.0257}$&${\bf 0.7399}_{0.0225}$&${\bf 0.6927}_{0.0220}$&${\bf 0.8243}_{0.0191}$&${\bf 0.8715}_{0.0106}$&${\bf 0.8448}_{0.0149}$&${\bf 0.8730}_{0.0108}$&${\bf 0.9019}_{0.0103}$&${\bf 0.8869}_{0.0102}$\\
    
    \hline\hline
    
    \hline
        \multicolumn{1}{|l|}{\bf Training Size}
        & \multicolumn{3}{c|}{\bf 150}
        & \multicolumn{3}{c|}{\bf 200}
        & \multicolumn{3}{c|}{\bf 250}
    \\ \cline{2-10}
    
    \multicolumn{1}{|c|}{}
        & {\bf Precision} & {\bf Recall} & {\bf F1}
        & {\bf Precision} & {\bf Recall} & {\bf F1}
        & {\bf Precision} & {\bf Recall} & {\bf F1}
    \\ \hline

    \textbf{CE}
        &${0.7761}_{0.0188}$&${0.8500}_{0.0110}$&${0.8094}_{0.0159}$&${0.8167}_{0.0149}$&${0.8916}_{0.0043}$&${0.8508}_{0.0103}$&${0.8317}_{0.0168}$&${0.9001}_{0.0037}$&${0.8616}_{0.0117}$\\

    \textbf{CRF}
        &${0.7056}_{0.0108}$&${0.7187}_{0.0124}$&${0.7099}_{0.0094}$&${0.7165}_{0.0076}$&${0.7254}_{0.0078}$&${0.7196}_{0.0053}$&${0.7293}_{0.0089}$&${0.7437}_{0.0081}$&${0.7351}_{0.0063}$\\
    
    \textbf{CE-2T}
        &${0.0951}_{0.0167}$&${0.3831}_{0.0082}$&${0.1336	}_{0.0031}$&${0.3626}_{0.0565}$&${0.4334}_{0.0186}$&${0.3291}_{0.0411}$&${0.3229}_{0.0553}$&${0.4139}_{0.0179}$&${0.2884}_{0.0384}$\\
    
    \textbf{AUC-2T}
        &$\underline{0.8718}_{0.0144}$&${\bf 0.9150}_{0.0069}$&$\underline{0.8921}_{0.0110}$&$\underline{0.8827}_{0.0066}$&${\bf 0.9180}_{0.0037}$&$\underline{0.8998}_{0.0048}$&$\underline{0.8875}_{0.0056}$&$\underline{0.9175}_{0.0058}$&$\underline{0.9020}_{0.0051}$\\
    
    \textbf{COMAUC-2T}
        &${\bf 0.8814}_{0.0091}$&$\underline{0.9110}_{0.0094}$&${\bf 0.8958}_{0.0091}$&${\bf 0.8867}_{0.0062}$&$\underline{0.9170}_{0.0035}$&${\bf 0.9014}_{0.0045}$&${\bf 0.8943}_{0.0039}$&${\bf 0.9198}_{0.0032}$&${\bf 0.9068}_{0.0034}$\\
    
    \hline\hline
    
    \hline
        \multicolumn{1}{|l|}{\bf Training Size}
        & \multicolumn{3}{c|}{\bf 300}
        & \multicolumn{3}{c|}{\bf 350}
        & \multicolumn{3}{c|}{\bf 400}
    \\ \cline{2-10}
    
    \multicolumn{1}{|c|}{}
        & {\bf Precision} & {\bf Recall} & {\bf F1}
        & {\bf Precision} & {\bf Recall} & {\bf F1}
        & {\bf Precision} & {\bf Recall} & {\bf F1}
    \\ \hline

    \textbf{CE}
        &${0.8503}_{0.0140}$&${0.9110}_{0.0040}$&${0.8778}_{0.0092}$&${0.8763}_{0.0069}$&${0.9192}_{0.0025}$&${0.8969}_{0.0043}$&${0.8880}_{0.0056}$&${0.9229}_{0.0022}$&${0.9049}_{0.0035}$\\
    \textbf{CRF}
        &${0.7460}_{0.0093}$&${0.7545}_{0.0070}$&${0.7489}_{0.0061}$&${0.7585}_{0.0073}$&${0.7696}_{0.0056}$&${0.7634}_{0.0053}$&${0.7698}_{0.0058}$&${0.7719}_{0.0063}$&${0.7700}_{0.0040}$\\
    
    \textbf{CE-2T}
        &${0.5172}_{0.0456}$&${0.5512}_{0.0322}$&${0.4990}_{0.0432}$&${0.6043}_{0.0332}$&${0.6695}_{0.0198}$&${0.6250}_{0.0315}$&${0.6298}_{0.0282}$&${0.6937}_{0.0192}$&${0.6529}_{0.0271}$\\
    
    \textbf{AUC-2T}
        &$\underline{0.8929}_{0.0084}$&$\underline{0.9185}_{0.0078}$&$\underline{0.9054}_{0.0081}$&${\bf 0.9091}_{0.0024}$&$\underline{0.9283}_{0.0024}$&$\underline{0.9185}_{0.0020}$&$\underline{0.9081}_{0.0032}$&${\bf 0.9324}_{0.0018}$&$\underline{0.9200}_{0.0019}$\\

    \textbf{COMAUC-2T}
        &${\bf 0.9021}_{0.0032}$&${\bf 0.9284}_{0.0027}$&${\bf 0.9150}_{0.0026}$&$\underline{0.9081}_{0.0023}$&${\bf 0.9298}_{0.0024}$&${\bf 0.9188}_{0.0020}$&${\bf 0.9092}_{0.0020}$&$\underline{0.9313}_{0.0024}$&${\bf 0.9200}_{0.0018}$\\
    
    \hline\hline
    
    \hline
        \multicolumn{1}{|l|}{\bf Training Size}
        & \multicolumn{3}{c|}{\bf 450}
        & \multicolumn{3}{c|}{\bf 500}
        & \multicolumn{3}{c|}{\bf 1000}
    \\ \cline{2-10}
    
    \multicolumn{1}{|c|}{}
        & {\bf Precision} & {\bf Recall} & {\bf F1}
        & {\bf Precision} & {\bf Recall} & {\bf F1}
        & {\bf Precision} & {\bf Recall} & {\bf F1}
    \\ \hline

    \textbf{CE}
        &${0.8926}_{0.0052}$&${0.9268}_{0.0022}$&${0.9092}_{0.0033}$&${0.9009}_{0.0046}$&${0.9292}_{0.0022}$&${0.9147}_{0.0031}$&$\underline{0.9294}_{0.0015}$&${\bf 0.9480}_{0.0013}$&${\bf 0.9386}_{0.0011}$\\

    \textbf{CRF}
        &${0.7673}_{0.0050}$&${0.7759}_{0.0084}$&${0.7705}_{0.0046}$&${0.7880}_{0.0048}$&${0.7834}_{0.0081}$&${0.7849}_{0.0050}$&${0.8336}_{0.0053}$&${0.8402}_{0.0052}$&${0.8365}_{0.0042}$\\
    
    \textbf{CE-2T}
        &${0.6633}_{0.0093}$&${0.7275}_{0.0107}$&${0.6928}_{0.0090}$&${0.6685}_{0.0077}$&${0.7386}_{0.0089}$&${0.7007}_{0.0070}$&${0.7552}_{0.0071}$&${0.8648}_{0.0066}$&${0.8060}_{0.0066}$\\

    \textbf{AUC-2T}
        &${\bf 0.9087}_{0.0033}$&$\underline{0.9311}_{0.0026}$&$\underline{0.9197}_{0.0023}$&${\bf 0.9141}_{0.0023}$&$\underline{0.9333}_{0.0024}$&${\bf 0.9235}_{0.0018}$&${0.9254}_{0.0014}$&${0.9450}_{0.0010}$&${0.9351}_{0.0009}$\\
    
    \textbf{COMAUC-2T}
        &$\underline{0.9086}_{0.0023}$&${\bf 0.9318}_{0.0020}$&${\bf 0.9200}_{0.0017}$&$\underline{0.9122}_{0.0025}$&${\bf 0.9343}_{0.0023}$&$\underline{0.9231}_{0.0019}$&${\bf 0.9303}_{0.0015}$&$\underline{0.9458}_{0.0010}$&$\underline{0.9379}_{0.0009}$\\
    
    \hline
    \end{tabular}
\end{adjustbox}
\caption{
Average performance taken from {30 random partitions} of different training set sizes. The non-parametric bootstrapped standard errors
from these experiments are under-scripted. 
The best performance for each setting is bold while the second best is underlined. The model architecture is ``transformers'' and the token embeddings is ``bert-base-cased'' \cite{devlin-etal-2019-bert}. 
}
\label{tab:CEvsAUC}
\end{table*}

\subsection{Ablation Studies}
\label{ablation-studies}
Under our AUC NER two-task setups, the hyperparameter $\lambda$ controls the trade-off between the two tasks (see Eq~\eqref{eq:AAAIAUCBothLoss}). 
Our intuitions tell us that $\lambda\gg 1$ since being in an NE is the dominant factor and unless 2 NEs run together, it is all that is needed.
We admit that choosing a good $\lambda$ for both AUC-2T and COMAUC-2T could be adhoc.
Thus, we greedily searched for the optimal $\lambda$ by training both setups multiple times with different partitions of 200 sentences and record their performance for different $\lambda$ values. 
The results are presented in Figure~\ref{fig:ablation_lamb} and the optimums are wide and flat.
Both Figure~\ref{fig:ablation_lamb_auc} and Figure~\ref{fig:ablation_lamb_comauc} suggest that  AUC-2T and COMAUC-2T reach their optimal performance at $\lambda=100$. 
Since the beginning-token prediction task is more imbalanced than the entity-token prediction task, we believe that $\lambda=100$ is large enough so that the gradients from $\text{AUC}_{\mathrm{M}}(\mathbf{w}_{\textbf{be}})$ is sufficiently represented when Eq~\eqref{eq:AAAIAUCBothLoss} is optimized.
For the rest of the paper, we present the results for AUC-2T and COMAUC-2T with $\lambda=100$ during training.

\subsection{Low-Resource Studies}
\begin{figure*}[t]
\centering
    \includegraphics[width=\textwidth]{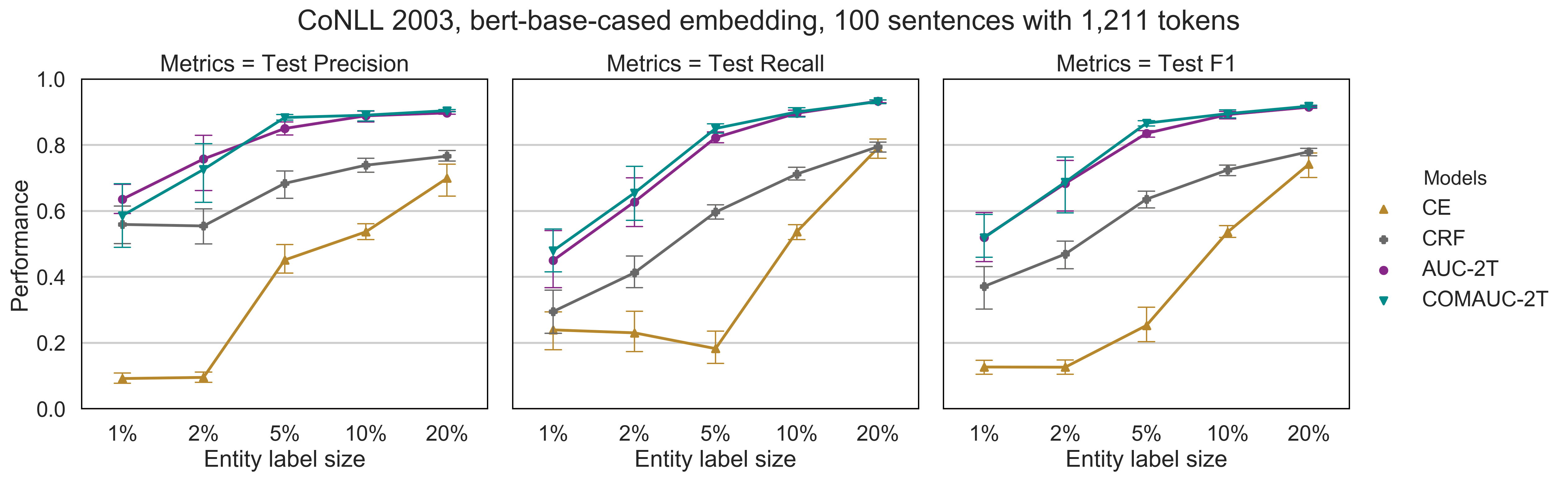}
    \caption{
    Learning curves of the average performance taken from 10 random data partitions, each partition with the size of 100 sentences ({\it i.e.,} $\approx$ 1,211 tokens).
    The original BIO label distribution for CoNLL 2003 is $11.5\%$, $5.2\%$, and $83.3\%$ respectively (see Table~\ref{data-description}), {\it i.e.}, original entity label size $=16.7\%$. 
    The entity label size in this graph represents the percentage of entity-tokens over the number of tokens for the training set $\mathcal{S}$. 
    The error bands indicate the $95\%$ confidence level of the scores. 
    }
    \label{fig:ablation_100sents_imb_CE_AUC}
\end{figure*}
Using the results from both Table~\ref{tab:CEvsAUC} in this section and Figure~5
in the appendix, we have the following observations:
\begin{itemize}
    \item \textbf{CE vs. AUC-2T}: 
    Under the extreme low-resource settings ({\it i.e.}, size of $\{20, 50\}$), AUC-2T  outperforms CE with a noticeable margin, with the average difference in the F1-performance  
    getting as big as $40\%$. 
    When the training set size gets bigger, AUC-2T still shows improvements compared to CE, most of the times with significant improvements at the 95\% level of confidence. 
    When the training set size is $1000$, CE outperforms AUC-2T on average; however, we argue that this is not significant and the difference is negligible. 
    Furthermore, except for the training set size  of $20$, 
    AUC-2T outputs more stable results compared to CE as indicated by their standard errors. 
    This comparison gives evidence that AUC-2T is a better objective function for the low-resource NER settings compared to the standard CE objective function.

    \item \textbf{CRF vs. AUC-2T}: 
    Albeit losing significantly across the performance metrics to AUC-2T under all settings, 
    CRF still produces quite stable results in some cases. 
    Nonetheless, given the results, we believe that AUC-2T is a better alternative to CRF under the low-resource NER settings.
    
    \item \textbf{CE-2T vs. AUC-2T}: 
    CE-2T is our last and weakest baseline; 
    thus, it is not surprising that it performs the worst.  
    This happens since while AUC-2T puts equal focus on both positive and negative-class tokens, CE, due to its nature of following the maximum likelihood principle, tends to favor the class with the majority presence during training.
    Although they should give similar results asymptotically, it is clear that all CE objective/loss functions suffer when the data is inherently imbalanced, as is the case of the NER task.
    Since we reformulate the original multi-class setting into the two-task setting, this amplifies the imbalanced data distribution issue, at least for the beginning-token prediction task.
    Thus, it is unsurprising that CE-2T gives poor performance and only recovers/improves when the size of $\mathcal{S}$  gets bigger.
    Comparing CE-2T with AUC-2T reveals that maximizing AUC scores contributes substantially to the 
    NER performance.
    \item \textbf{COMAUC-2T vs. AUC-2T}: 
    Although the F1-score of COMAUC-2T is higher on average than that of AUC-2T under almost all settings, the difference between the two approaches is not significant, particularly when the size of $\mathcal{S}$ is large.
    Since COMAUC-2T alternates between the standard multi-class cross entropy loss function 
    and the AUC two-task loss function during training, we surmise the difference is due to 
    1) better learned feature representations ({\it i.e.}, embeddings, transformers) from minimizing the standard CE loss function as theoretically proven by \citet{yuan2021compositional}, and
    2) higher training time complexity as COMAUC-2T time complexity $=$ AUC-2T time complexity $+$ CE time complexity. Since COMAUC-2T performs on par with AUC-2T, it consequently outperforms all baselines in most low-resource NER settings.
\end{itemize}
We can observe similar result patterns to those mentioned above in the sub-section ``low-resource studies'' found in the supplementary material.
These results are collected from different data domains (general, disease, species), embeddings (bert-base-cased, facebook/bart-base, biobert-base-cased-v1.1), and model architectures (transformers, seq2seq).
Consequently, they strengthen the evidence that our method is domain, embedding, and model agnostic.

\begin{table*}[!t]
\setlength{\tabcolsep}{.5pt}
\centering
\small
\textbf{Predicting PER (5.4\% of tokens) as the entity type}
\begin{adjustbox}{max width=\textwidth}
\begin{tabular}{|l|ccc|ccc|ccc|}
    \hline
    \multicolumn{1}{|l|}{\bf Training Size}
        & \multicolumn{3}{c|}{\bf 50}
        & \multicolumn{3}{c|}{\bf 100}
        & \multicolumn{3}{c|}{\bf 200}
    \\ \cline{2-10}
    
    \multicolumn{1}{|c|}{}
        & {\bf Precision} & {\bf Recall} & {\bf F1}
        & {\bf Precision} & {\bf Recall} & {\bf F1}
        & {\bf Precision} & {\bf Recall} & {\bf F1}
    \\ \hline 
    
    \textbf{DL}&        
        ${0.0177}_{0.0021}$&${0.0062}_{0.0006}$&${0.0089}_{0.0008}$&${0.0188}_{0.0014}$&${0.0070}_{0.0005}$&${0.0101}_{0.0007}$&${0.0219}_{0.0022}$&${0.0068}_{0.0010}$&${0.0102}_{0.0012}$\\

    \textbf{AUC-2T}
        &${\bf 0.8102}_{0.0161}$&${\bf 0.8199}_{0.0274}$&${\bf 0.8131}_{0.0188}$&$\underline{0.8640}_{0.0086}$&${\bf 0.8975}_{0.0112}$&${\bf 0.8796}_{0.0058}$&${\bf 0.8824}_{0.0109}$&$\underline{0.8955}_{0.0090}$&$\underline{0.8884}_{0.0076}$\\

    \textbf{COMAUC-2T}
        &$\underline{ 0.8037}_{0.0153}$&$\underline{0.8134}_{0.0253}$&$\underline{0.8068}_{0.0175}$&${\bf 0.8786}_{0.0117}$&$\underline{0.8461}_{0.0170}$&$\underline{0.8606}_{0.0095}$&$\underline{0.8727}_{0.0059}$&${\bf 0.9067}_{0.0104}$&${\bf 0.8890}_{0.0060}$\\

    \hline\hline
    
    \hline
        \multicolumn{1}{|l|}{\bf Training Size}
        & \multicolumn{3}{c|}{\bf 400}
        & \multicolumn{3}{c|}{\bf 500}
        & \multicolumn{3}{c|}{\bf 1000}
    \\ \cline{2-10}
    
    \multicolumn{1}{|c|}{}
        & {\bf Precision} & {\bf Recall} & {\bf F1}
        & {\bf Precision} & {\bf Recall} & {\bf F1}
        & {\bf Precision} & {\bf Recall} & {\bf F1}
    \\ \hline

    \textbf{DL}&        
        ${0.0636}_{0.0062}$&${0.0154}_{0.0012}$&${0.0246}_{0.0019}$&${0.1859}_{0.0101}$&${0.0421}_{0.0025}$&${0.0677}_{0.0033}$&${0.7893}_{0.0105}$&${0.5505}_{0.0082}$&${0.6486}_{0.0090}$\\

    \textbf{AUC-2T}
        &$\underline{0.9021}_{0.0054}$&${\bf 0.9144}_{0.0086}$&$\underline{0.9078	}_{0.0043}$&${\bf 0.9140}_{0.0074}$&${\bf 0.9249}_{0.0073}$&${\bf 0.9191}_{0.0054}$&$\underline{0.9181}_{0.0057}$&${\bf 0.9354}_{0.0032}$&$\underline{0.9266}_{0.0039}$\\
    
    \textbf{COMAUC-2T}
        &${\bf 0.9094}_{0.0044}$&$\underline{0.9116}_{0.0045}$&${\bf 0.9105}_{0.0038}$&$\underline{0.9115}_{0.0064}$&$\underline{0.9213}_{0.0081}$&$\underline{0.9162}_{0.0061}$&${\bf 0.9296}_{0.0064}$&$\underline{0.9341}_{0.0046}$&${\bf 0.9317}_{0.0039}$\\
    
    \hline
    \end{tabular}
\end{adjustbox}

\textbf{Predicting LOC (4.1\% of tokens) as the entity type}

\begin{adjustbox}{max width=\textwidth}
\begin{tabular}{|l|ccc|ccc|ccc|}
    \hline
        \multicolumn{1}{|l|}{\bf Training Size}
        & \multicolumn{3}{c|}{\bf 50}
        & \multicolumn{3}{c|}{\bf 100}
        & \multicolumn{3}{c|}{\bf 200}
    \\ \cline{2-10}
    \multicolumn{1}{|c|}{}
        & {\bf Precision} & {\bf Recall} & {\bf F1}
        & {\bf Precision} & {\bf Recall} & {\bf F1}
        & {\bf Precision} & {\bf Recall} & {\bf F1}
    \\ \hline 

    \textbf{DL}&        
        ${0.0460}_{0.0041}$&${0.0229}_{0.0018}$&${0.0297}_{0.0021}$&${0.0332}_{0.0030}$&${0.0178}_{0.0012}$&${0.0226}_{0.0013}$&${0.0349}_{0.0035}$&${0.0176}_{0.0008}$&${0.0230}_{0.0013}$\\

    \textbf{AUC-2T}
        &$\underline{0.5225}_{0.0195}$&$\underline{0.6674}_{0.0436}$&$\underline{0.5779}_{0.0241}$&$\underline{0.6203}_{0.0273}$&${\bf 0.6959}_{0.0232}$&$\underline{0.6512}_{0.0182}$&$\underline{0.6547}_{0.0293}$&${\bf 0.8153}_{0.0149}$&$\underline{0.7205}_{0.0148}$\\
    
    \textbf{COMAUC-2T}
        &${\bf 0.5828}_{0.0311}$&${\bf 0.6766}_{0.0334}$&${\bf 0.6180}_{0.0218}$&${\bf 0.6468}_{0.0262}$&$\underline{0.6886}_{0.0300}$&$\underline{0.6629}_{0.0218}$&${\bf 0.7044}_{0.0273}$&$\underline{0.7784}_{0.0159}$&${\bf 0.7341}_{0.0125}$\\
    
    \hline\hline
    
    \hline
        \multicolumn{1}{|l|}{\bf Training Size}
        & \multicolumn{3}{c|}{\bf 400}
        & \multicolumn{3}{c|}{\bf 500}
        & \multicolumn{3}{c|}{\bf 1000}
    \\ \cline{2-10}
    
    \multicolumn{1}{|c|}{}
        & {\bf Precision} & {\bf Recall} & {\bf F1}
        & {\bf Precision} & {\bf Recall} & {\bf F1}
        & {\bf Precision} & {\bf Recall} & {\bf F1}
    \\ \hline
    \textbf{DL}&        
        ${0.0469}_{0.0068}$&${0.0224}_{0.0017}$&${0.0288}_{0.0024}$&${0.1209}_{0.0204}$&${0.0282}_{0.0025}$&${0.0451}_{0.0045}$&${0.5953}_{0.0142}$&${0.5656}_{0.0162}$&${0.5766}_{0.0064}$\\

    \textbf{AUC-2T}
        &${\bf 0.7594}_{0.0180}$&${\bf 0.8306}_{0.0467}$&${\bf 0.7908}_{0.0084}$&${\bf 0.7737}_{0.0163}$&${\bf 0.8269}_{0.0132}$&${\bf 0.7979}_{0.0108}$&$\underline{0.8291}_{0.0172}$&${\bf 0.8818}_{0.0169}$&$\underline{0.8538}_{0.0103}$\\
    
    \textbf{COMAUC-2T}
        &$\underline{0.7492}_{0.0224}$&$\underline{0.8267}_{0.0076}$&$\underline{0.7835}_{0.0114}$&$\underline{0.7641}_{0.0178}$&$\underline{0.8220}_{0.0082}$&$\underline{0.7910}_{0.0118}$&${\bf 0.8467}_{0.0180}$&$\underline{0.8706}_{0.0059}$&${\bf 0.8573}_{0.0100}$\\
    
    \hline
    \end{tabular}
\end{adjustbox}

\textbf{Predicting ORG (4.9\% of tokens) as the entity type}

\begin{adjustbox}{max width=\textwidth}
\begin{tabular}{|l|ccc|ccc|ccc|}
    \hline
        \multicolumn{1}{|l|}{\bf Training Size}
        & \multicolumn{3}{c|}{\bf 50}
        & \multicolumn{3}{c|}{\bf 100}
        & \multicolumn{3}{c|}{\bf 200}
    \\ \cline{2-10}
    
    \multicolumn{1}{|c|}{}
        & {\bf Precision} & {\bf Recall} & {\bf F1}
        & {\bf Precision} & {\bf Recall} & {\bf F1}
        & {\bf Precision} & {\bf Recall} & {\bf F1}
    \\ \hline 

    \textbf{DL}&        
        ${0.0394}_{0.0108}$&${0.0056}_{0.0011}$&${0.0093}_{0.0020}$&${0.0113}_{0.0025}$&${0.0047}_{0.0004}$&${0.0061}_{0.0005}$&${0.0103}_{0.0044}$&${0.0064}_{0.0020}$&${0.0056}_{0.0018}$\\

    \textbf{AUC-2T}
        &${\bf 0.4536}_{0.0377}$&${\bf 0.3155}_{0.0471}$&${\bf 0.3401}_{0.0320}$&${\bf 0.4720}_{0.0288}$&${\bf 0.4250}_{0.0572}$&${\bf 0.4288}_{0.0401}$&${\bf 0.6147}_{0.0262}$&${\bf 0.5128}_{0.0248}$&${\bf 0.5566}_{0.0227}$\\
    
    \textbf{COMAUC-2T}
        &$\underline{0.3921}_{0.0253}$&$\underline{0.2906}_{0.0459}$&$\underline{0.3113}_{0.0336}$&$\underline{0.4509}_{0.0334}$&$\underline{0.3764}_{0.0514}$&$\underline{0.3907}_{0.0385}$&$\underline{0.5982}_{0.0181}$&$\underline{0.4845}_{0.0249}$&$\underline{0.5312}_{0.0190}$\\
    
    \hline\hline
    
    \hline
        \multicolumn{1}{|l|}{\bf Training Size}
        & \multicolumn{3}{c|}{\bf 400}
        & \multicolumn{3}{c|}{\bf 500}
        & \multicolumn{3}{c|}{\bf 1000}
    \\ \cline{2-10}
    
    \multicolumn{1}{|c|}{}
        & {\bf Precision} & {\bf Recall} & {\bf F1}
        & {\bf Precision} & {\bf Recall} & {\bf F1}
        & {\bf Precision} & {\bf Recall} & {\bf F1}
    \\ \hline
    
    \textbf{DL}&        
        ${0.0031}_{0.0005}$&${0.0034}_{0.0008}$&${0.0026}_{0.0002}$&${0.0117}_{0.0038}$&${0.0033}_{0.0009}$&${0.0044}_{0.0012}$&${0.3212}_{0.0381}$&${0.2059}_{0.0265}$&${0.2236}_{0.0242}$\\

    \textbf{AUC-2T}
        &$\underline{0.6799}_{0.0255}$&${\bf 0.6265}_{0.0386}$&${\bf 0.6469}_{0.0292}$&$\underline{0.6959}_{0.0134}$&${\bf 0.6518}_{0.0344}$&${\bf 0.6671}_{0.0208}$&$\underline{0.7608}_{0.0160}$&$\underline{0.7498}_{0.0322}$&$\underline{0.7533}_{0.0240}$\\
    
    \textbf{COMAUC-2T}
        &${\bf 0.6906	}_{0.0186}$&$\underline{0.6105}_{0.0368}$&$\underline{0.6442}_{0.0259}$&${\bf 0.7129}_{0.0135}$&$\underline{0.6217}_{0.0358}$&$\underline{0.6552}_{0.0223}$&${\bf 0.7669}_{0.0185}$&${\bf 0.7601}_{0.0303}$&${\bf 0.7612}_{0.0220}$\\
    
    \hline
    \end{tabular}
\end{adjustbox}

\textbf{Predicting MISC (2.3\% of tokens) as the entity type}

\begin{adjustbox}{max width=\textwidth}
\begin{tabular}{|l|ccc|ccc|ccc|}
    \hline
        \multicolumn{1}{|l|}{\bf Training Size}
        & \multicolumn{3}{c|}{\bf 50}
        & \multicolumn{3}{c|}{\bf 100}
        & \multicolumn{3}{c|}{\bf 200}
    \\ \cline{2-10}
    
    \multicolumn{1}{|c|}{}
        & {\bf Precision} & {\bf Recall} & {\bf F1}
        & {\bf Precision} & {\bf Recall} & {\bf F1}
        & {\bf Precision} & {\bf Recall} & {\bf F1}
    \\ \hline 

    \textbf{DL}&        
        ${0.0021}_{0.0010}$&${0.0013}_{0.0005}$&${0.0010}_{0.0003}$&${0.0018}_{0.0004}$&${0.0038}_{0.0013}$&${0.0017}_{0.0002}$&${0.0033}_{0.0011}$&${0.0031}_{0.0007}$&${0.0027}_{0.0006}$\\

    \textbf{AUC-2T}
        &$\underline{0.4257}_{0.0776}$&$\underline{0.2969}_{0.0503}$&$\underline{0.3292}_{0.0572}$&$\underline{ 0.4234}_{0.0746}$&$\underline{0.3718}_{0.0439}$&$\underline{0.3816}_{0.0527}$&$\underline{0.5638}_{0.0436}$&${\bf 0.5456}_{0.0319}$&${\bf 0.5486}_{0.0318}$\\
    
    \textbf{COMAUC-2T}
        &${\bf 0.4710}_{0.0771}$&${\bf 0.3311}_{0.0438}$&${\bf 0.3805}_{0.0550}$&${\bf 0.4351}_{0.0714}$&${\bf 0.3772}_{0.0437}$&${\bf 0.3941}_{0.0522}$&${\bf 0.5707}_{0.0499}$&$\underline{0.5395}_{0.0271}$&$\underline{0.5462}_{0.0355}$\\
    
    \hline\hline
    
    \hline
        \multicolumn{1}{|l|}{\bf Training Size}
        & \multicolumn{3}{c|}{\bf 400}
        & \multicolumn{3}{c|}{\bf 500}
        & \multicolumn{3}{c|}{\bf 1000}
    \\ \cline{2-10}
    
    \multicolumn{1}{|c|}{}
        & {\bf Precision} & {\bf Recall} & {\bf F1}
        & {\bf Precision} & {\bf Recall} & {\bf F1}
        & {\bf Precision} & {\bf Recall} & {\bf F1}
    \\ \hline
    
    \textbf{DL}&        
        ${0.0007}_{0.0001}$&${0.0059}_{0.0016}$&${0.0011}_{0.0001}$&${0.0009}_{0.0001}$&${0.0064}_{0.0015}$&${0.0014}_{0.0001}$&${0.0007}_{0.0001}$&${0.0118}_{0.0001}$&${0.0014}_{0.0001}$\\
        
    \textbf{AUC-2T}
        &$\underline{0.6341}_{0.0304}$&${\bf 0.6060}_{0.027}$&$\underline{0.6184}_{0.0269}$&$\underline{0.6468	}_{0.0294}$&$\underline{0.6091}_{0.0306}$&$\underline{0.6260}_{0.0289}$&${\bf 0.7385	}_{0.0110}$&$\underline{0.6830}_{0.0088}$&${\bf 0.7091}_{0.0075}$\\
    
    \textbf{COMAUC-2T}
        &${\bf 0.6525}_{0.0268}$&$\underline{0.6047}_{0.0177}$&${\bf 0.6264}_{0.0203}$&${\bf 0.6744}_{0.0316}$&${\bf 0.6155}_{0.0232}$&${\bf 0.6426}_{0.0264}$&$\underline{0.7308}_{0.0118}$&${\bf 0.6869}_{0.0082}$&$\underline{0.7073}_{0.0064}$\\
    
    \hline
    \end{tabular}
\end{adjustbox}
\caption{
Average performance taken from {10 random partitions} of different training set sizes. The non-parametric bootstrapped standard errors
from these experiments are under-scripted. 
The best performance for each setting is bold while the second best is underlined. The model architecture is ``transformers'' and the token embeddings is ``bert-base-cased'' \cite{devlin-etal-2019-bert}. 
}
\label{tab:dicelosscomp}
\end{table*}

\subsection{Imbalanced Data Distribution Studies}
As previously mentioned, we developed an imbalanced data generator for the NER task. 
This generator serves to simulate the scenarios where the data distribution for $\mathcal{S}$ is different from that of $\mathcal{S}_{\text{test}}$, testing the robustness of both the baselines and our method.
We present the main result in Figure~\ref{fig:ablation_100sents_imb_CE_AUC}.
From this result, we have the following observations:
\begin{itemize}
    \item \textbf{CE vs. AUC-2T}: 
    We found that with the extreme imbalanced training sets ({\it i.e.}, entity label size of $1$ and $2\%$), CE performs extremely poorly even when the training size is reasonably adequate. 
    The F1-score drops even lower than training with 20 sentences as indicated in Table~\ref{tab:CEvsAUC}.
    Although the F1-score for AUC-2T also dropped given these settings, the drop is not as significant as CE. 
    Additionally, we observe that the performance of AUC-2T under these extreme settings is still on par with that of CE when training with the normal training set $\mathcal{S}$ (Table~\ref{tab:CEvsAUC}).
    \item \textbf{CRF vs. AUC-2T}: 
    Although CRF performance drops under the extreme imbalanced settings, the decline is not as severe as that of CE, showing the robustness of CRF loss function.
    Additionally, both AUC-2T and CRF improve with more entities in $\mathcal{S}$ at a similar rate.
    However, as AUC excels under the imbalanced settings, the performance gap between AUC-2T and CRF is still significant.
    \item \textbf{COMAUC-2T vs. AUC-2T}: Both approaches show similar F1-performance across different entity label size.
    As AUC-2T has shown its robustness compared to other loss/objective functions, COMAUC-2T consequently can be considered as a better alternative to its baselines.
    \item Lastly, the results also indicate
    that 
    {having more NEs in $\mathcal{S}$ will lead to higher F1-scores on $\mathcal{S}_{\text{test}}$ for all approaches, regardless of the data distribution differences between $\mathcal{S}$ and $\mathcal{S}_{\text{test}}$}. We surmise this happens as more NEs means more information for training the  model parameters. This is consistent with the results from \citet{nguyen-etal-2022-hardness}.
\end{itemize}
Note that we excluded CE-2T from our results and discussions as it already exhibits poor performance under the standard imbalance setting.
We also provide experimental results for different training-set sizes, domains, embeddings, and model architectures in sub-section ``imbalanced data distribution studies'' in the supplementary material. 
From these results, we can observe similar result patterns with those in Figure~\ref{fig:ablation_100sents_imb_CE_AUC}, indicating the robustness of our approach.

\subsection{Comparison with Other NER Imbalanced Losses}
Lastly, we compare our low-resource NER solutions to Dice Loss \cite{li-etal-2020-dice}, which is based on the Sørensen–Dice coefficient \cite{sorensen1948method} to alleviate the influences of negative examples in the NER problem. The corpus we used for our experiments is CoNLL 2003 as it is the corpus used by \citet{li-etal-2020-dice} in their paper to present the findings. As Dice Loss (DL) attempts to find both the entity tokens and their entity type, we modify the original codes~\footnote{\text{https://github.com/ShannonAI/dice\_loss\_for\_NLP}} so that both our methods and Dice Loss only focus on a specific entity type, say LOC, {\it i.e.}, we train the model to predict the BIO tag for only the LOC tokens. Please note that since we only predict for the a single type of entities ({\it e.g.,} LOC), the percentage of entity tokens is drastically lower than 16.7\% (see Table~\ref{data-description}) , making the distribution more imbalanced and the problem harder to solve. We present the experimental results in Table~\ref{tab:dicelosscomp} and provide the following observations:

\begin{itemize}
    \item \textbf{AUC-2T vs. COMAUC-2T}: 
    Our two methods still exhibit consistently strong performance under this new experimental setting. The difference in term of performance (precision, recall and f1) between the two methods, with a 95\% confident level, are not significant under all cases, which aligns with our findings from previous sections.
    \item \textbf{AUC-2T vs. DL}:
    Under the low-resource settings, our AUC-2T outperforms Dice Loss significantly. We presume this happens since while Dice Loss shows great performance when all training data are available \cite{li-etal-2020-dice}, the low-resource settings exacerbate the challenge of the NER tasks. For entity type that appears sparingly in the corpus ({\it e.g.,} MISC), Dice Loss struggles even with 1000 training sentences as there is little learning signals for its algorithm to work with. Although Dice Loss demonstrates great improvement for some entity types once the training pool is large enough, we believe that both of our methods should be the preferred alternative under the low-resource NER settings. As AUC-2T outperforms Dice Loss significantly, our COMAUC-2T also displays a strong performance against Dice Loss.
\end{itemize}

\section{Conclusions}
In this paper, we studied an effective solution to the low-resource and the data imbalance issues
that widely exist in many NER/BioNER tasks.
To tackle these two issues, we have 
first reformulated the conventional NER task as a two-task learning problem,
which consists of two binary classifiers predicting if a word is a part of an NE and if the word is the 
start of an NE respectively. We then 
adapted the idea of AUC maximization to develop a new NER loss based on the reformulation.
Extensive experiments on different datasets with different scenarios, which mimic the low-resource and
the data imbalance issues, demonstrated that the new AUC-based loss function performs substantially better 
than the commonly used CE and CRF, regardless of the underlying NER models, embeddings or domains that we used.
Although the proposed AUC optimization approach works quite well for NER, there still exist some limitations,
including the inconsistency in the prediction that arises from the two-task reformulation of NER.
We believe that 
the AUC optimization for the multi-class problem \cite{yang2021learning} could be an alternative, which is subject to future work.
\bibliography{aaai23}

\begin{thebibliography}{45}
\providecommand{\natexlab}[1]{#1}

\bibitem[{Banerjee et~al.(2019)Banerjee, Chakraborty, Tripathi, Gupta, and
  Kumar}]{banerjee2019information}
Banerjee, P.~S.; Chakraborty, B.; Tripathi, D.; Gupta, H.; and Kumar, S.~S.
  2019.
\newblock A information retrieval based on question and answering and NER for
  unstructured information without using SQL.
\newblock \emph{Wireless Personal Communications}, 108(3): 1909--1931.

\bibitem[{Chen et~al.(2021)Chen, Aguilar, Neves, and Solorio}]{chen2021data}
Chen, S.; Aguilar, G.; Neves, L.; and Solorio, T. 2021.
\newblock Data Augmentation for Cross-Domain Named Entity Recognition.
\newblock In \emph{Proceedings of the 2021 Conference on Empirical Methods in
  Natural Language Processing}, 5346--5356. Online and Punta Cana, Dominican
  Republic: Association for Computational Linguistics.

\bibitem[{Cl{\'e}men{\c{c}}on, Lugosi, and
  Vayatis(2008)}]{clemenccon2008ranking}
Cl{\'e}men{\c{c}}on, S.; Lugosi, G.; and Vayatis, N. 2008.
\newblock Ranking and empirical minimization of U-statistics.
\newblock \emph{The Annals of Statistics}, 36(2): 844--874.

\bibitem[{Devlin et~al.(2019)Devlin, Chang, Lee, and
  Toutanova}]{devlin-etal-2019-bert}
Devlin, J.; Chang, M.-W.; Lee, K.; and Toutanova, K. 2019.
\newblock {BERT}: Pre-training of Deep Bidirectional Transformers for Language
  Understanding.
\newblock In \emph{Proceedings of the 2019 Conference of the North {A}merican
  Chapter of the Association for Computational Linguistics: Human Language
  Technologies, Volume 1 (Long and Short Papers)}, 4171--4186. Minneapolis,
  Minnesota: Association for Computational Linguistics.

\bibitem[{Do\u{g}an, Leaman, and Lu(2014)}]{10.5555/2598938.2599127}
Do\u{g}an, R.~I.; Leaman, R.; and Lu, Z. 2014.
\newblock Special Report: {NCBI} Disease Corpus: A Resource for Disease Name
  Recognition and Concept Normalization.
\newblock \emph{J. of Biomedical Informatics}, 47: 1–10.

\bibitem[{Etzioni et~al.(2005)Etzioni, Cafarella, Downey, Popescu, Shaked,
  Soderland, Weld, and Yates}]{etzioni2005unsupervised}
Etzioni, O.; Cafarella, M.; Downey, D.; Popescu, A.-M.; Shaked, T.; Soderland,
  S.; Weld, D.~S.; and Yates, A. 2005.
\newblock Unsupervised named-entity extraction from the web: An experimental
  study.
\newblock \emph{Artificial intelligence}, 165(1): 91--134.

\bibitem[{Freund et~al.(2003)Freund, Iyer, Schapire, and
  Singer}]{freund2003efficient}
Freund, Y.; Iyer, R.; Schapire, R.~E.; and Singer, Y. 2003.
\newblock An efficient boosting algorithm for combining preferences.
\newblock \emph{Journal of machine learning research}, 4(Nov): 933--969.

\bibitem[{Gao et~al.(2013)Gao, Jin, Zhu, and Zhou}]{gao2013one}
Gao, W.; Jin, R.; Zhu, S.; and Zhou, Z.-H. 2013.
\newblock One-pass AUC optimization.
\newblock In \emph{International conference on machine learning}, 906--914.
  PMLR.

\bibitem[{Gerner, Nenadic, and Bergman(2010)}]{article10}
Gerner, M.; Nenadic, G.; and Bergman, C. 2010.
\newblock LINNAEUS: A species name identification system for biomedical
  literature.
\newblock \emph{BMC bioinformatics}, 11: 85.

\bibitem[{Giorgi and Bader(2019)}]{10.1093/bioinformatics/btz504}
Giorgi, J.~M.; and Bader, G.~D. 2019.
\newblock {Towards reliable named entity recognition in the biomedical domain}.
\newblock \emph{Bioinformatics}, 36(1): 280--286.

\bibitem[{Hanley and McNeil(1982)}]{hanley1982meaning}
Hanley, J.~A.; and McNeil, B.~J. 1982.
\newblock The meaning and use of the area under a receiver operating
  characteristic (ROC) curve.
\newblock \emph{Radiology}, 143(1): 29--36.

\bibitem[{Huang, Xu, and Yu(2015)}]{huang2015bidirectional}
Huang, Z.; Xu, W.; and Yu, K. 2015.
\newblock Bidirectional LSTM-CRF Models for Sequence Tagging.
\newblock arXiv:1508.01991.

\bibitem[{Kotlowski, Dembczynski, and
  Huellermeier(2011)}]{kotlowski2011bipartite}
Kotlowski, W.; Dembczynski, K.; and Huellermeier, E. 2011.
\newblock Bipartite ranking through minimization of univariate loss.
\newblock In \emph{ICML}.

\bibitem[{Kripke(1980)}]{Kripke1980-KRINAN}
Kripke, S. 1980.
\newblock \emph{Naming and Necessity}.
\newblock Harvard University Press.

\bibitem[{Lample et~al.(2016)Lample, Ballesteros, Subramanian, Kawakami, and
  Dyer}]{DBLP:journals/corr/LampleBSKD16}
Lample, G.; Ballesteros, M.; Subramanian, S.; Kawakami, K.; and Dyer, C. 2016.
\newblock Neural Architectures for Named Entity Recognition.
\newblock In \emph{Proceedings of the 2016 Conference of the North {A}merican
  Chapter of the Association for Computational Linguistics: Human Language
  Technologies}, 260--270. San Diego, California: Association for Computational
  Linguistics.

\bibitem[{Lee et~al.(2019)Lee, Yoon, Kim, Kim, Kim, So, and
  Kang}]{10.1093/bioinformatics/btz682}
Lee, J.; Yoon, W.; Kim, S.; Kim, D.; Kim, S.; So, C.~H.; and Kang, J. 2019.
\newblock {{BioBERT}: a pre-trained biomedical language representation model
  for biomedical text mining}.
\newblock \emph{Bioinformatics}, 36(4): 1234--1240.

\bibitem[{Lewis et~al.(2020)Lewis, Liu, Goyal, Ghazvininejad, Mohamed, Levy,
  Stoyanov, and Zettlemoyer}]{lewis2019bart}
Lewis, M.; Liu, Y.; Goyal, N.; Ghazvininejad, M.; Mohamed, A.; Levy, O.;
  Stoyanov, V.; and Zettlemoyer, L. 2020.
\newblock {BART}: Denoising Sequence-to-Sequence Pre-training for Natural
  Language Generation, Translation, and Comprehension.
\newblock In \emph{Proceedings of the 58th Annual Meeting of the Association
  for Computational Linguistics}, 7871--7880. Online: Association for
  Computational Linguistics.

\bibitem[{Li, Shang, and Shao(2020)}]{10.1145/3366423.3380127}
Li, J.; Shang, S.; and Shao, L. 2020.
\newblock MetaNER: Named Entity Recognition with Meta-Learning.
\newblock In \emph{Proceedings of The Web Conference 2020}, WWW ’20,
  429–440. New York, NY, USA: Association for Computing Machinery.
\newblock ISBN 9781450370233.

\bibitem[{{Li} et~al.(2020){Li}, {Sun}, {Han}, and
  {Li}}]{DBLP:journals/corr/abs-1812-09449}
{Li}, J.; {Sun}, A.; {Han}, J.; and {Li}, C. 2020.
\newblock A Survey on Deep Learning for Named Entity Recognition.
\newblock \emph{IEEE Transactions on Knowledge and Data Engineering}, 1--1.

\bibitem[{Li et~al.(2020)Li, Sun, Meng, Liang, Wu, and Li}]{li-etal-2020-dice}
Li, X.; Sun, X.; Meng, Y.; Liang, J.; Wu, F.; and Li, J. 2020.
\newblock Dice Loss for Data-imbalanced {NLP} Tasks.
\newblock In \emph{Proceedings of the 58th Annual Meeting of the Association
  for Computational Linguistics}, 465--476. Online: Association for
  Computational Linguistics.

\bibitem[{Liu et~al.(2020)Liu, Yuan, Ying, and Yang}]{liu2019stochastic}
Liu, M.; Yuan, Z.; Ying, Y.; and Yang, T. 2020.
\newblock Stochastic AUC Maximization with Deep Neural Networks.
\newblock In \emph{International Conference on Learning Representations}.

\bibitem[{Liu et~al.(2018)Liu, Zhang, Chen, Wang, and Yang}]{liu2018fast}
Liu, M.; Zhang, X.; Chen, Z.; Wang, X.; and Yang, T. 2018.
\newblock Fast stochastic AUC maximization with $ o (1/n) $-convergence rate.
\newblock In \emph{International Conference on Machine Learning}, 3189--3197.
  PMLR.

\bibitem[{Nakayama(2018)}]{seqeval}
Nakayama, H. 2018.
\newblock {seqeval}: A Python framework for sequence labeling evaluation.
\newblock \url{https://github.com/chakki-works/seqeval}.
\newblock Accessed: 2022-07-15.

\bibitem[{Natole, Ying, and Lyu(2018)}]{natole2018stochastic}
Natole, M.; Ying, Y.; and Lyu, S. 2018.
\newblock Stochastic proximal algorithms for AUC maximization.
\newblock In \emph{International Conference on Machine Learning}, 3710--3719.
  PMLR.

\bibitem[{Nguyen et~al.(2022)Nguyen, Du, Buntine, Chen, and
  Beare}]{nguyen-etal-2022-hardness}
Nguyen, N.~D.; Du, L.; Buntine, W.; Chen, C.; and Beare, R. 2022.
\newblock Hardness-guided domain adaptation to recognise biomedical named
  entities under low-resource scenarios.
\newblock In \emph{Proceedings of the 2022 Conference on Empirical Methods in
  Natural Language Processing}, 4063--4071. Abu Dhabi, United Arab Emirates:
  Association for Computational Linguistics.

\bibitem[{Pafilis et~al.(2013)Pafilis, Frankild, Fanini, Faulwetter, Pavloudi,
  Vasileiadou, Arvanitidis, and Jensen}]{10.1371/journal.pone.0065390}
Pafilis, E.; Frankild, S.~P.; Fanini, L.; Faulwetter, S.; Pavloudi, C.;
  Vasileiadou, A.; Arvanitidis, C.; and Jensen, L.~J. 2013.
\newblock The SPECIES and ORGANISMS Resources for Fast and Accurate
  Identification of Taxonomic Names in Text.
\newblock \emph{PLOS ONE}, 8(6): 1--6.

\bibitem[{Patel et~al.(2005)Patel, Patel, Arocha, and
  Zhang}]{Patel2005ThinkingAR}
Patel, V.~L.; Patel, N.; Arocha, J.~F.; and Zhang, J. 2005.
\newblock Thinking and Reasoning in Medicine.
\newblock \emph{The Cambridge handbook of thinking and reasoning}, 14:
  727--750.

\bibitem[{Peng et~al.(2021)Peng, Li, Li, Shayandeh, Liden, and
  Gao}]{peng2020soloist}
Peng, B.; Li, C.; Li, J.; Shayandeh, S.; Liden, L.; and Gao, J. 2021.
\newblock Soloist: Building Task Bots at Scale with Transfer Learning and
  Machine Teaching.
\newblock \emph{Transactions of the Association for Computational Linguistics},
  9: 807--824.

\bibitem[{Rahimi, Li, and Cohn(2019)}]{rahimi2019massively}
Rahimi, A.; Li, Y.; and Cohn, T. 2019.
\newblock Massively Multilingual Transfer for {NER}.
\newblock In \emph{Proceedings of the 57th Annual Meeting of the Association
  for Computational Linguistics}, 151--164. Florence, Italy: Association for
  Computational Linguistics.

\bibitem[{Ritter, Etzioni, and Clark(2012)}]{ritter2012open}
Ritter, A.; Etzioni, O.; and Clark, S. 2012.
\newblock Open domain event extraction from {T}witter.
\newblock In \emph{Proceedings of the 18th ACM SIGKDD international conference
  on Knowledge discovery and data mining}, 1104--1112.

\bibitem[{Sorensen(1948)}]{sorensen1948method}
Sorensen, T.~A. 1948.
\newblock A method of establishing groups of equal amplitude in plant sociology
  based on similarity of species content and its application to analyses of the
  vegetation on Danish commons.
\newblock \emph{Biol. Skar.}, 5: 1--34.

\bibitem[{Tan, Du, and Buntine(2021)}]{NEURIPS2021_5a7b238b}
Tan, W.; Du, L.; and Buntine, W. 2021.
\newblock Diversity Enhanced Active Learning with Strictly Proper Scoring
  Rules.
\newblock In Ranzato, M.; Beygelzimer, A.; Dauphin, Y.; Liang, P.; and Vaughan,
  J.~W., eds., \emph{Advances in Neural Information Processing Systems},
  volume~34, 10906--10918. Curran Associates, Inc.

\bibitem[{Tjong Kim~Sang and
  De~Meulder(2003)}]{tjong-kim-sang-de-meulder-2003-introduction}
Tjong Kim~Sang, E.~F.; and De~Meulder, F. 2003.
\newblock Introduction to the {C}o{NLL}-2003 Shared Task: Language-Independent
  Named Entity Recognition.
\newblock In \emph{Proceedings of the Seventh Conference on Natural Language
  Learning at {HLT}-{NAACL} 2003}, 142--147.

\bibitem[{Weischedel et~al.(2014)Weischedel, Pradhan, Ramshaw, Palmer, Xue,
  Marcus, Taylor, Greenberg, Hovy, Belvin et~al.}]{weischedel2011ontonotes}
Weischedel, R.; Pradhan, S.; Ramshaw, L.; Palmer, M.; Xue, N.; Marcus, M.;
  Taylor, A.; Greenberg, C.; Hovy, E.; Belvin, R.; et~al. 2014.
\newblock Ontonotes release 5.0.
\newblock \emph{LDC2011T03, Philadelphia, Penn.: Linguistic Data Consortium}.

\bibitem[{Xu et~al.(2019)Xu, Yang, Kang, Wang, and Liu}]{XU2019122}
Xu, K.; Yang, Z.; Kang, P.; Wang, Q.; and Liu, W. 2019.
\newblock Document-level attention-based {BiLSTM-CRF} incorporating disease
  dictionary for disease named entity recognition.
\newblock \emph{Computers in Biology and Medicine}, 108: 122 -- 132.

\bibitem[{Xu et~al.(2018)Xu, Zhou, Hao, and Liu}]{10.1007/978-3-319-64861-3_33}
Xu, K.; Zhou, Z.; Hao, T.; and Liu, W. 2018.
\newblock A Bidirectional LSTM and Conditional Random Fields Approach to
  Medical Named Entity Recognition.
\newblock In Hassanien, A.~E.; Shaalan, K.; Gaber, T.; and Tolba, M.~F., eds.,
  \emph{Proceedings of the International Conference on Advanced Intelligent
  Systems and Informatics 2017}, 355--365. Cham: Springer International
  Publishing.
\newblock ISBN 978-3-319-64861-3.

\bibitem[{Yadav and Bethard(2018)}]{yadav2019survey}
Yadav, V.; and Bethard, S. 2018.
\newblock A Survey on Recent Advances in Named Entity Recognition from Deep
  Learning models.
\newblock In \emph{Proceedings of the 27th International Conference on
  Computational Linguistics}, 2145--2158. Santa Fe, New Mexico, USA:
  Association for Computational Linguistics.

\bibitem[{Yang et~al.(2021)Yang, Xu, Bao, Cao, and Huang}]{yang2021learning}
Yang, Z.; Xu, Q.; Bao, S.; Cao, X.; and Huang, Q. 2021.
\newblock Learning with Multiclass AUC: Theory and Algorithms.
\newblock \emph{IEEE Transactions on Pattern Analysis and Machine
  Intelligence}.

\bibitem[{Ying, Wen, and Lyu(2016)}]{ying2016stochastic}
Ying, Y.; Wen, L.; and Lyu, S. 2016.
\newblock Stochastic online auc maximization.
\newblock \emph{Advances in neural information processing systems}, 29.

\bibitem[{Yuan et~al.(2021{\natexlab{a}})Yuan, Guo, Chawla, and
  Yang}]{yuan2021compositional}
Yuan, Z.; Guo, Z.; Chawla, N.; and Yang, T. 2021{\natexlab{a}}.
\newblock Compositional Training for End-to-End Deep AUC Maximization.
\newblock In \emph{International Conference on Learning Representations}.

\bibitem[{Yuan et~al.(2022)Yuan, Qiu, Li, Zhu, Guo, Hu, Wang, Qi, Zhong, and
  Yang}]{libauc2022}
Yuan, Z.; Qiu, Z.-H.; Li, G.; Zhu, D.; Guo, Z.; Hu, Q.; Wang, B.; Qi, Q.;
  Zhong, Y.; and Yang, T. 2022.
\newblock {LibAUC}: A Deep Learning Library for X-risk Optimization.
\newblock \url{https://https://libauc.org/}.
\newblock Accessed: 2022-07-11.

\bibitem[{Yuan et~al.(2021{\natexlab{b}})Yuan, Yan, Sonka, and
  Yang}]{yuan2020large}
Yuan, Z.; Yan, Y.; Sonka, M.; and Yang, T. 2021{\natexlab{b}}.
\newblock Large-scale Robust Deep AUC Maximization: A New Surrogate Loss and
  Empirical Studies on Medical Image Classification.
\newblock In \emph{2021 IEEE/CVF International Conference on Computer Vision
  (ICCV)}, 3020--3029. Los Alamitos, CA, USA: IEEE Computer Society.

\bibitem[{Zhao et~al.(2011)Zhao, Hoi, Jin, and Yang}]{zhao2011online}
Zhao, P.; Hoi, S. C.~H.; Jin, R.; and Yang, T. 2011.
\newblock Online AUC Maximization.
\newblock In \emph{Proceedings of the 28th International Conference on
  International Conference on Machine Learning}, ICML'11, 233–240. Madison,
  WI, USA: Omnipress.
\newblock ISBN 9781450306195.

\bibitem[{Zhou et~al.(2022)Zhou, Li, He, Bing, Cambria, Si, and
  Miao}]{zhou2022melm}
Zhou, R.; Li, X.; He, R.; Bing, L.; Cambria, E.; Si, L.; and Miao, C. 2022.
\newblock MELM: Data Augmentation with Masked Entity Language Modeling for
  Low-Resource NER.
\newblock In \emph{Proceedings of the 60th Annual Meeting of the Association
  for Computational Linguistics (Volume 1: Long Papers)}, 2251--2262.

\bibitem[{Zuva and Zuva(2012)}]{zuva2012evaluation}
Zuva, K.; and Zuva, T. 2012.
\newblock Evaluation of information retrieval systems.
\newblock \emph{AIRCC's International Journal of Computer Science and
  Information Technology}, 4(3): 35--43.

\end{thebibliography}
\end{document}